\documentclass[preprint,12pt]{elsarticle}

\usepackage{amssymb}

\usepackage{lineno}

\usepackage{wrapfig}
\usepackage{amsmath}
\usepackage{multirow}
\usepackage{color}
\usepackage{subcaption}
\usepackage{algorithm}
\usepackage{algorithmic}
\usepackage{booktabs}
\usepackage{mathrsfs}
\usepackage{makecell}
\usepackage{graphicx}
\usepackage{arydshln}
\usepackage[colorlinks]{hyperref}

\usepackage{diagbox}

\journal{arxiv.org}

\begin{document}

\begin{frontmatter}

    \title{Rethinking the Sample Relations for Few-Shot Classification}

    \author[label1]{Guowei Yin}

\author[label1,label2]{Sheng Huang\corref{cor1}}

    \author[label3,label4,label5]{Luwen Huangfu}


    \author[label1]{Yi Zhang}

    \author[label1,label2]{Xiaohong Zhang}

    \cortext[cor1]{Corresponding author}
    

    \address[label1]{School of Big Data and Software Engineering, Chongqing University,
        No.55 Daxuecheng South Rd.,
        Shapingba,
        401331,
        Chongqing,
        China}

 \address[label2]{Ministry of Education Key Laboratory of Dependable Service Computing in Cyber Physical Society, Chongqing University,
	No.174 Shazhengjie,
	Shapingba,
	400044,
	Chongqing,
	China}

    \address[label3]{Fowler College of Business, San Diego State University,
    San Diego,
    California,
    92182,
    USA}

    \address[label4]{Center for Human Dynamics in the Mobile Age (HDMA), San Diego State University,
    San Diego,
    California,
    92182,
    USA}

    \address[label5]{AI4Business Lab, San Diego State University,
    San Diego,
    California,
    92182,
    USA}

    \begin{abstract}
        Feature quality is paramount for classification performance, particularly in few-shot scenarios. Contrastive learning, a widely adopted technique for enhancing feature quality, leverages sample relations to extract intrinsic features that capture semantic information and has achieved remarkable success in Few-Shot Learning (FSL). Nevertheless, current few-shot contrastive learning approaches often overlook the semantic similarity discrepancies at different granularities when employing the same modeling approach for different sample relations, which limits the potential of few-shot contrastive learning. In this paper, we introduce a straightforward yet effective contrastive learning approach, Multi-Grained Relation Contrastive Learning (MGRCL), as a pre-training feature learning model to boost few-shot learning by meticulously modeling sample relations at different granularities. MGRCL categorizes sample relations into three types: intra-sample relation of the same sample under different transformations, intra-class relation of homogenous samples, and inter-class relation of inhomogeneous samples. In MGRCL, we design Transformation Consistency Learning (TCL) to ensure the rigorous semantic consistency of a sample under different transformations by aligning predictions of input pairs. Furthermore, to preserve discriminative information, we employ Class Contrastive Learning (CCL) to ensure that a sample is always closer to its homogenous samples than its inhomogeneous ones, as homogenous samples share similar semantic content while inhomogeneous samples have different semantic content. Our method is assessed across four popular FSL benchmarks, showing that such a simple pre-training feature learning method surpasses a majority of leading FSL methods. Moreover, our method can be incorporated into other FSL methods as the pre-trained model and help them obtain significant performance gains.
    \end{abstract}


\end{frontmatter}

\section{Introduction} \label{intro}
Nowadays, deep learning has achieved significant advancements in various artificial intelligence tasks \cite{krizhevsky2017imagenet, yolo, FasterRcnn}, which are heavily dependent on the presence of ample labeled data. However, in practical scenarios, obtaining large-scale labeled data is often challenging due to the scarcity of relevant images and the substantial expense associated with manual annotation. To address this problem, the concept of Few-shot learning (FSL) has emerged. The aim of FSL is to train a model using abundant labeled data from base classes and then apply this acquired knowledge to novel classes that only have a limited number of labeled samples per class (e.g., 1 or 5).

To tackle the FSL problem, numerous approaches have been developed. Many of these approaches adopt meta-learning to solve FSL by either designing optimal algorithms \cite{MAML, optimization, lee2019meta} or learning good metrics \cite{ProtoNet, vinyals2016matching, DeepEMD, zhu2023light}. These methods simulate the FSL tasks during the training phase and attempt to train a base model that can swiftly adjust to new tasks. For instance, \cite{zhu2023light} proposes a novel light transformer-based global information enhanced metric-learning classification model to obtain better embedding for FSL. Additionally, based on the intuition of adding extra samples to mitigate the issue of limited data, many data-augmentation-based methods \cite{IDeMe-Net, DualTriNet, AFHN, STVAE} have been proposed. They enhance sample diversity by synthesizing additional samples. Such as, STVAE \cite{STVAE} introduces a generative FSL approach that exploits the complementarity of semantic and visual prior information to synthesize features for novel classes. However, these approaches always involve complex training phases or need to add many extra samples in the testing phase, which are computationally expensive. Recent works \cite{dhillon2019baseline, chencloser, RFS} have shown that fully supervised pre-training of a model on the entire base dataset, followed by freezing the model as a feature extractor in the meta-testing phase can achieve competitive performance to these sophisticated methods. This reveals the vital importance of feature learning in FSL.

To obtain a better feature extractor, numerous researchers focus on feature learning of FSL and propose various impressive works \cite{RFS, IER, PAL, HandCrafted, Spatial, lee2020self, IEPT, ESPT}. Among them, some approaches utilize self-supervised tasks to improve the feature extraction capability of networks \cite{lee2020self, IEPT, ESPT}. For example, ESPT\cite{ESPT} maximizes the local spatial relationship consistency between the original episode and the transformed one. And many other approaches employ contrastive learning as an auxiliary task \cite{IER, PAL, Spatial} because contrastive learning can mitigate the problem that the network only learns the most discriminative features for classifying base classes through cross-entropy loss while overlooking the acquisition of certain sub-discriminative features, which reduces the transferability of the network to novel classes. For example, IER \cite{IER} leverages an unsupervised contrastive loss to constrain the invariance of images under different transformations. PAL \cite{PAL} uses supervised contrastive learning \cite{SupCon} for the initial training of a teacher model, which is subsequently leveraged to supply soft labels for the student model.

These aforementioned methods employing contrastive learning as an auxiliary task achieve good performance, but they may not fully exploit the potential of sample relations by directly using unsupervised or supervised contrastive learning methods. Unsupervised contrastive learning \cite{SimCLR, MoCo} operates by considering different transformed versions of the same sample as positive pairs, while treating distinct samples as negative pairs, irrespective of their respective class labels. Although this approach effectively increases the separation between samples from different classes, it also inadvertently drives apart diverse samples belonging to the same class, which is detrimental to feature learning to some extent. Supervised contrastive learning \cite{SupCon} overcomes the aforementioned issue by ensuring that samples from identical classes are more closely aligned than those from disparate classes. But it treats transformed versions of a sample and other homogenous samples equally when modeling sample relations. This strategy is not appropriate, since a sample always has exactly the same semantic content with its transformed versions while only sharing similar semantic content with its homogenous samples. In other words, samples should be closer to their transformed versions than to their homogenous samples in the learned feature space. 

\begin{figure}[h]
    \centering
    \vspace{-5pt}
    \includegraphics[scale=1.2]{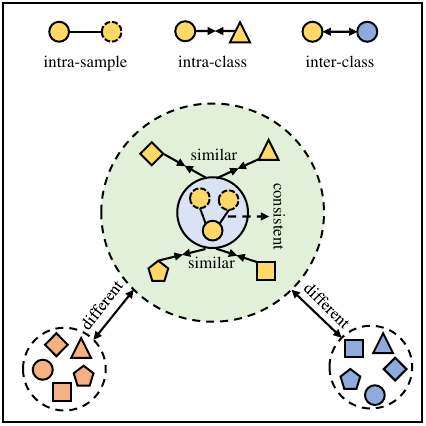}        
    \caption{In this figure, shapes and colors represent different samples and classes respectively. Different transformations of the same sample are represented by the same color and shape. The sample relations contain three types: intra-sample relation of the same sample under different transformations, intra-class relation of homogenous samples, and inter-class relation of inhomogeneous samples. Our approach enforces different transformations to be consistent in semantic content, homogenous samples to be similar, and inhomogeneous samples to be different. Unsupervised contrastive learning only constrains the intra-sample relation, and supervised contrastive learning treats the intra-sample relation as equal to the intra-class relation.}
    \label{image_SampleRelation}
\vspace{-10pt}
\end{figure}

To solve these issues, we rethink different relations of samples and present a novel Multi-Grained Relation Contrastive Learning approach (MGRCL) for few-shot learning. MGRCL categorizes the sample relations into three types with different granularities: intra-sample relation of the same sample under different transformations, intra-class relation of homogenous samples, and inter-class relation of inhomogeneous samples, as shown in Fig. \ref{image_SampleRelation}. In MGRCL, the first type is the intra-sample relation, we constrain it by Transformation Consistency Learning (TCL). TCL ensures consistency in label outputs for a sample and its transformed versions by aligning the predicted label distributions, which can guarantee that different transformations of one sample maintain the same semantic content. The second type is the intra-class relation of homogenous samples, which cannot be constrained in this way because their semantic content differs from transformed versions. Neglecting the semantic discrepancy among homogenous samples and mapping them into an identical position in feature space will easily lead to model collapse. To avoid this, we employ Class Contrastive Learning (CCL) to regulate both this type and the third type, which is the inter-class relation of inhomogeneous samples, in a relative way. CCL primarily focuses on ensuring discrimination and does not require complete consistency among features of homogenous samples. Instead, it emphasizes distinguishing between identical and different classes based on the relative distances of their features. To validate the efficacy of our approach, we extensively conduct experiments on four benchmarks, which include three general few-shot classification datasets: miniImageNet \cite{vinyals2016matching}, tieredImageNet \cite{ren2018meta}, CIFAR-FS \cite{bertinetto2018meta}, and a fine-grained few-shot classification dataset: CUB-200-2011 \cite{wah2011caltech}.
 
The principal contributions of this paper are as follows:
\begin{itemize}
\item We rethink the sample relations in few-shot learning and categorize them into three distinct types based on the different granularities. By exploiting different sample relations, we present a novel Multi-Grained Relation Contrastive Learning approach (MGRCL) for few-shot learning.

\item To constrain the semantic similarity of sample relations at different granularities, we design two simple yet highly effective components, namely Transformation Consistency Learning and Class Contrastive Learning.

\item Experimental results on four benchmarks suggest that our approach exhibits a commendable performance compared with recent methods. Furthermore, our approach can provide a good pre-trained model for these two-stage meta-learning methods and generative methods to improve their performance. 
\end{itemize}

\section{Related Work} \label{related}
\subsection{Few-shot Learning}
The aim of few-shot learning (FSL) is to identify new classes using a minimal number of examples, typically one or five. Methods for FSL generally fall into three primary categories: meta-learning-based, data-augmentation-based, and feature-learning-based.

Methods based on meta-learning form a significant subset of FSL techniques. These can be further categorized into optimization-based and metric-based. In particular, optimization-based methods \cite{MAML, optimization, lee2019meta} aim to learn how to swiftly adjust model parameters to fit new tasks. Metric-based methods \cite{ProtoNet, vinyals2016matching, DeepEMD} project images into a metric space and perform classification by calculating the distance between the sample and known classified samples. To mimic the meta-testing phase, these methods require sampling a large number of FSL tasks during the pre-training phase, and some of them necessitate retraining the model when the FSL task changes (from 1-shot to 5-shot), which is computationally costly.

Data-augmentation-based methods \cite{IDeMe-Net, DualTriNet, AFHN, STVAE} are another branch of FSL methods, which use generative models or other methods to synthesize additional training samples for FSL tasks. For instance, IDeMe-Net \cite{IDeMe-Net} proposes an image deformation framework that generates extra images by linearly fusing the patches of probe and gallery images. STVAE \cite{STVAE} introduces a generative FSL approach that exploits the complementarity of semantic and visual prior information to synthesize features for novel classes. Most of them need to train an extra generative model and add plenty of samples in the testing phase.

Recently, lots of feature-learning-based methods \cite{RFS, IER, PAL, HandCrafted, Spatial} have emerged as effective solutions for FSL. These methods directly use the entire base dataset to train a feature extractor in a fully supervised manner or add additional auxiliary self-supervised tasks. For instance, RFS \cite{RFS} learns feature embeddings by training a network on the entire base dataset. During the testing phase, it freezes the network to extract features from images and then adds a logistic regression classifier to perform FSL. Other works like \cite{IER, HandCrafted} build upon RFS and further enhance the capacity of the pre-trained network to extract features by adding self-supervised tasks. Following these works, our proposed method also uses the entire base dataset to train a model and does not need to sample FSL tasks during training or add extra generative models.

\subsection{Contrastive Learning}
As a feature learning method, contrastive learning has recently gained popularity due to its ability to derive meaningful representations from images. Many contrastive learning methods have been developed for image classification \cite{SimCLR, MoCo, SupCon, guo2023contrastive} and other vision tasks \cite{wang2022arco, zhao2023dual, zhao2023unsupervised} in recent years. Among these methods, unsupervised contrastive learning methods likely SimCLR \cite{SimCLR} and MoCo \cite{MoCo} treat different transformed versions of the same sample as positive pairs while treating distinct samples as negative pairs. However, a limitation of these approaches is that while they push samples of different classes further apart, these methods also inadvertently push different samples of the same class further apart, which hinders feature learning in supervised tasks. To address this issue, the supervised contrastive learning method, SupCon \cite{SupCon} has been proposed, which not only treats different transformed versions of the same sample as positive pairs but also treats the samples from the same classes as positive pairs. However, one drawback of SupCon is that it treats transformed versions of a sample as same as other homogenous samples.

To enhance the ability of the model to generalize and extract features, lots of FSL approaches employ contrastive learning to model the sample relations, such as \cite{IER, PAL, Spatial}. However, both unsupervised contrastive learning and supervised contrastive learning have some issues mentioned before, they do not take into account the modeling of sample relations with different granularity separately. So, in this paper, we rethink sample relations and introduce a method to model the sample relations at different granularities.

\section{Methodology} \label{methodology}
\subsection{Problem Formulation}
\label{sec_ProblemFormulation}
In few-shot learning (FSL), the dataset can be denoted as $\mathcal{D} = \{\mathcal{D}_{base}, \mathcal{D}_{novel}\}$. $\mathcal{D}_{base}$ is the base dataset with $\mathcal{C}_{base}$ classes, and $\mathcal{D}_{novel}$ is the novel dataset with $\mathcal{C}_{novel}$ classes, where $\mathcal{C}_{novel}$ is disjoint from $\mathcal{C}_{base}$. FSL strives to obtain a well-generalized feature extractor using $\mathcal{D}_{base}$ and can achieve good performance in $\mathcal{D}_{novel}$ during the testing phase.

In FSL, $\mathcal{D}_{base}$ is used to train a well-generalized model in the pre-training phase. The testing phase contains lots of FSL tasks drawn from $\mathcal{D}_{novel}$, and each FSL task can be regarded as a \emph{N}-way \emph{K}-shot classification problem, where \emph{N} represents the number of categories, \emph{K}  represents the number of labeled samples per category. Usually, $\emph{N} = 5$ and $\emph{K} = 1 \:\text{or}\: 5$.  Each task $\mathcal{T}$ includes a support set $\mathcal{S}_\mathcal{T}$ and a query set $\mathcal{Q}_\mathcal{T}$,

\begin{equation}
  \mathcal{T} = (\mathcal{S}_\mathcal{T}, \mathcal{Q}_\mathcal{T}).
\end{equation}

\noindent Here, $\mathcal{S}_\mathcal{T}$ contains \emph{K} labeled samples from each of \emph{N} categories, and $\mathcal{Q}_\mathcal{T}$ contains \emph{Q} samples from the same \emph{N} classes. $\mathcal{S}_\mathcal{T}$ and $\mathcal{Q}_\mathcal{T}$ are disjoint. In the testing phase, the model is trained using $\mathcal{S}_\mathcal{T}$, while $\mathcal{Q}_\mathcal{T}$ is employed to evaluate the performance.

\begin{figure}[h]
\centerline{\includegraphics[scale=0.55]{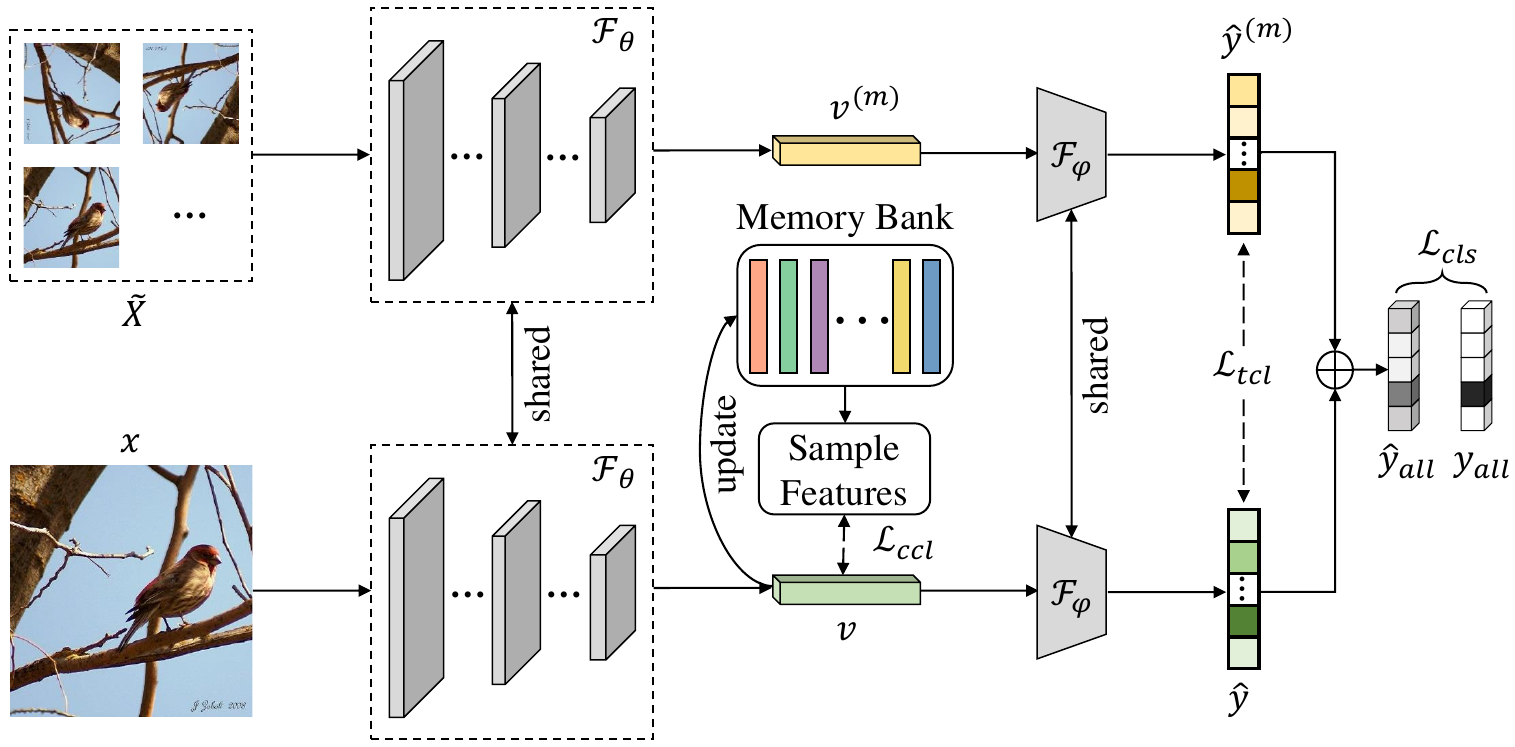}}
\caption{Network architecture of our model, which contains a CNN backbone $\mathcal{F}_{\theta}$ and a classifier $\mathcal{F}_{\varphi}$. Our proposed Transformation Consistency Learning and Class Contrastive Learning constrain different sample relations at the label level and feature level, respectively. In this figure, $v$ and $v^{(m)}$ represent the feature embeddings of the original image and its $m$-th transformed version. $\bigoplus$ is a concatenation operator for the predicted output $\widehat{y}$ of the original image and the predicted outputs $\{\widehat{y}^{(1)}, ..., \widehat{y}^{(M)}$\} of $M$ transformations. $\widehat{y}_{all}$ and $y_{all}$ are the predicted outputs and the ground truths of the original image and its transformations. And memory bank is used to store the feature embeddings.}
\label{fig_model}
\vspace{-10pt}
\end{figure}

\subsection{Overview}
Recently, lots of feature-learning-based FSL approaches utilize contrastive learning to explore the sample relations \cite{IER, PAL, Spatial} and achieve a promising performance. However, these approaches typically employ unsupervised or supervised contrastive learning directly, without considering the semantic differences in sample relations at different granularities and modeling them in detail.

To solve this issue, we rethink the sample relations and categorize them into three types according to the different granularities: intra-sample relation of the same sample under different transformations, intra-class relation of homogenous samples, and inter-class relation of inhomogeneous samples. Then we present a novel Multi-Grained Relation Contrastive Learning approach (MGRCL) for FSL to model the different sample relations. As illustrated in Fig. \ref{fig_model}, MGRCL contains three main components: the base learner, the Transformation Consistency Learning (TCL), and the Class Contrastive Learning (CCL). In particular, the base learner is a neural network trained by a general image classification task. TCL is designed to constrain the different transformations of the same sample to have consistent semantic content. And CCL is used to constrain that homogenous samples have similar semantic content and nonhomogeneous samples have different semantic content. Next, we will provide a more detailed explanation of each component.

\subsection{The Base Learner}
As depicted in Fig. \ref{fig_model}, the base learner, denoted as $\mathcal{F}_{\theta}$ with parameters $\theta$, is employed for the extraction of feature embeddings. Let $(x, y)\in \mathcal{D}_{base}$ denote an image and its corresponding label sampled from $\mathcal{D}_{base}$. The feature vector $v$ of an image $x$ can be obtained by $\mathcal{F}_{\theta}$: $v=\mathcal{F}_{\theta}(x)$. Then, a classifier $\mathcal{F}_{\varphi}$ with parameters $\varphi$ is employed to get the predicted confidence scores $p$ by projecting the feature vector $v$ into the label space: $p=\mathcal{F}_{\varphi}(v)$. Finally, we can derive the predicted label $\widehat{y}$ by applying the softmax operator on $p$: $\widehat{y}=\text{Softmax}(p)$. The parameters $\theta$, $\varphi$ of the base learner are optimized jointly by minimizing classification loss $\mathcal{L}_{cls}$ on the entire base dataset $\mathcal{D}_{base}$,

\begin{equation}
  \mathcal{L}_{cls} = - \frac{1}{|\mathcal{D}_{base}|}\sum_{\{x,y\}\in \mathcal{D}_{base}}y\log\widehat{y}.
\end{equation}

To avoid overfitting on the training set, many methods \cite{IER, PAL, SSLforFSL} incorporate transformed samples and predict which image transformation is performed during training. Following these methods, we also add a self-supervised module (SS) with a Multilayer Perceptron (MLP). Consider $\widetilde{X}=\{\widetilde{x}^{(1)}, ..., \widetilde{x}^{(M)}\}$ as the collection of transformed versions of a single image, where $M$ denotes the total number of transformed samples and $\widetilde{x}^{(m)}$ signifies the $m$-th transformed version of the image. $\widetilde{X}$ can be obtained by applying a series of transformations on the image, such as cropping, resizing, rotation, etc. The transformed images $\widetilde{X}$, together with the original image $x$, are simultaneously input into the model for both classification and self-supervised tasks. The objective of the self-supervised task is to identify the transformation applied to the image,

\begin{equation}
  \mathcal{L}_{ss} = - \frac{1}{|\mathcal{D}_{base}|}\frac{1}{M+1}\sum_{x \in \mathcal{D}_{base}}\sum_{m=0}^{M}s^{(m)}\log\widehat{s}^{(m)},
\end{equation}

\noindent where $\widehat{s}^{(m)}$ and $s^{(m)}$ represent the predicted output and the ground truth of $m$-th transformed version in self-supervised task. $s^{(m)}$ is the one-hot vector of $m$ and $s^{(0)}$ is the self-supervised label of the original image $x$. Additionally, classification loss can be redefined as,

\begin{equation}
  \mathcal{L}_{cls} = - \frac{1}{|\mathcal{D}_{base}|}\frac{1}{M+1}\sum_{x \in \mathcal{D}_{base}}\sum_{m=0}^{M}y^{(m)}\log\widehat{y}^{(m)},
\end{equation}

\noindent where $\widehat{y}^{(m)}$ denote the predicted output and $y^{(m)}$ denote the ground truth in the classification task.

Then the loss of the base learner can be written by

\begin{equation}
  \mathcal{L}_{base} = \mathcal{L}_{cls} + \mathcal{L}_{ss}.
\end{equation}

\subsection{Transformation Consistency Learning}
The different transformations of one sample should have the same semantic content as their original version since these images contain exactly the same objects and backgrounds. To accomplish this objective, we design a Transformation Consistency Learning (TCL) component to constrain the intra-sample relation of the same sample under different transformations. The label output can represent the semantic content of the sample because it reflects the predicted probability of the sample in each category. Therefore, to ensure that the transformations of one sample have consistent semantic content, we constrain them to have the same label outputs.

In the base learner, we feed the transformed versions of one sample into the network along with its original version, and calculate the TCL loss on their predicted label outputs. Here, we use Jensen-Shannon Divergence \cite{JS1, JS2} as TCL loss to constrain the intra-sample relation,

\begin{equation}
  \mathcal{L}_{tcl} = \frac{1}{|\mathcal{D}_{base}|}\sum_{x \in \mathcal{D}_{base}}\frac{1}{M}\sum_{m=1}^{M}JS(\widehat{y}_{\tau_1}, \widehat{y}_{\tau_1}^{(m)}),
\end{equation}

\noindent where $\widehat{y}_{\tau_1}$ and $\widehat{y}_{\tau_1}^{(m)}$ are the smoothed label outputs of the original image and the $m$-th transformed image, respectively. They are obtained by,

\begin{equation}
  \widehat{y}_{\tau_1} = \text{Softmax}(p/\tau_1),
\end{equation}

\noindent here $p = \mathcal{F}_{\varphi}(\mathcal{F}_{\theta}(x))$, $\tau_1$ is a temperature parameter, we set it to $4.0$ in experiments. The reason why we use the smoothed label outputs is that they can provide more information about the differences in probability distributions. And the outputs of different transformations need to be consistent not only for the category with the maximum prediction probability but also for all other categories to ensure they have exactly the same semantic content.

\subsection{Class Contrastive Learning}
To enforce the intra-class relation and the inter-class relation, we use Class Contrastive Learning (CCL) to maximize the similarity of feature embeddings of homogenous samples while minimizing the similarity of feature embeddings of inhomogeneous samples. Therefore, for each image, we need another homogenous sample and several inhomogeneous samples. To achieve this goal and accelerate training, we employ a memory bank \cite{wu2018unsupervised} to store and sample feature vectors. The memory bank holds the feature embeddings for all images in the dataset, enabling the model to sample a diverse set of features from a wider range of images, rather than being restricted to the current mini-batch. For each mini-batch, a feature embedding is sampled from the memory bank for each class of images, then CCL loss can be defined as,

\begin{equation}
  \mathcal{L}_{ccl} = \frac{1}{|\mathcal{D}_{base}|}\sum_{x \in \mathcal{D}_{base}}-\log \frac{exp(\frac{cos(v, v^\prime)}{\tau_2})}{\sum_{i=1}^{|\mathcal{C}_{base}|}{exp(\frac{cos(v, v_i)}{\tau_2})}},
\end{equation}

\noindent where $|\mathcal{C}_{base}|$ is the number of base classes, $v$, $v^\prime$ are the feature embedding of a sample and its homogenous sample, $v_i$ represents the feature embedding of a sample from $i$-th class. Here $v^\prime$ and $v_i$ are sampled from the memory bank. $cos(\cdot)$ is the cosine similarity. And $\tau_2$ is a temperature parameter, we set it to $0.1$ following \cite{SimCLR, SupCon}. Moreover, the memory bank is updated by,

\begin{equation}
  v_{k} = r\times v_k + (1 - r)\times v_q,
\end{equation}

\noindent where $v_q$ and $v_k$ represent the feature embedding of an image obtained in the current mini-batch and the same image stored in the memory bank, $r$ is used to adjust the updating speed of the memory bank, and we set it to 0.99 following \cite{IER}. During the training phase, the memory bank is completely updated with each epoch.

\subsection{Overall Optimization Objective}
In summary, by incorporating both two components which constrain sample relations at different granularities, the total loss can be written as,

\begin{equation}
  \mathcal{L}_{total} = \mathcal{L}_{base} + \alpha \cdot \mathcal{L}_{tcl} + \beta \cdot \mathcal{L}_{ccl},
\end{equation}

\noindent where $\alpha$ and $\beta$ are the hyper-parameters for balancing the different losses. 

\subsection{Few-shot Evaluation}
As depicted in Section \ref{sec_ProblemFormulation}, during the testing phase, the performance of our model is assessed by tackling abundant FSL tasks that are drawn from $\mathcal{D}_{novel}$. For every task $\mathcal{T}$, we keep our model parameters constant and employ the feature extractor $\mathcal{F}_{\theta}$ to derive the feature representations for $\mathcal{S}_\mathcal{T}$ and $\mathcal{Q}_\mathcal{T}$. Then, we employ a logistic regression classifier to classify samples of $\mathcal{Q}_\mathcal{T}$, which is trained on the feature representations of $\mathcal{S}_\mathcal{T}$.

\section{Experiments} \label{experiment}
\subsection{Datasets and Implementation Details}
\subsubsection{Datasets} 
We conduct experiments on four popular FSL benchmarks, which include three general datasets: miniImageNet \cite{vinyals2016matching}, tieredImageNet \cite{ren2018meta}, CIFAR-FS \cite{bertinetto2018meta}, and a fine-grained dataset: CUB-200-2011 (CUB) \cite{wah2011caltech}. For all datasets, we keep the splitting protocol as same as \cite{RENet}. And in our experiments, for miniImageNet, tieredImageNet, and CUB, the image size is 84$\times$84, while for CIFAR-FS, the image size is 32$\times$32.

\subsubsection{Backbone Architectures}
Following previous works \cite{RFS, DeepEMD, IER}, we adopt ResNet-12 as our backbone. Transformation Consistency Learning (TCL) is performed on the label outputs and Class Contrastive Learning (CCL) is performed on the features after global pooling. These techniques do not require additional network layers. And we add a self-supervised learning module with an MLP consisting of two fully connected layers, one batch-normalization layer, and an activation function.

\subsubsection{Optimization Setup} 
For all experiments, we employ the SGD optimizer with a momentum of 0.9 and a weight decay of $5e^{-4}$. The learning rate is initially set to 0.05 and is subsequently reduced by a factor of 0.1. For tieredImageNet, we train 60 epochs, and the learning rate decays after epochs 30, 40, and 50 respectively. For other datasets, we train 80 epochs, and the learning rate decays after epoch 60 and epoch 70. For the CIFAR-FS dataset, we set the batch size to 64, and for other datasets, we set the batch size to 32. Regarding the hyper-parameters, we have assigned the following values: $\alpha=1.0$, $\beta=0.1$, $\tau_1=4.0$, $\tau_2=0.1$.

\subsubsection{Data Augmentation}
To alleviate overfitting problems and implement our Transformation Consistency Learning (TCL), we add some augmented samples to train the feature extractor. The data augmentation contains three scaling transformations, three rotation transformations, one random erasing, one graying, and one Sobel edge detection.

\subsubsection{Evaluation Protocol}
For all datasets, we sample 2,000 few-shot classification tasks and compute the mean classification accuracy along with a 95\% confidence interval to assess the performance of our method. Each task can be regarded as a \emph{N}-way \emph{K}-shot classification problem, as described in Section \ref{sec_ProblemFormulation}. For each task $\mathcal{T} = (\mathcal{S}_\mathcal{T}, \mathcal{Q}_\mathcal{T})$, 5 categories are selected randomly from the dataset $\mathcal{D}_{novel}$. The support set $\mathcal{S}_\mathcal{T}$ contains either 1 or 5 labeled samples per category selected, depending on whether the task is a 1-shot or 5-shot scenario. Meanwhile, the query set $\mathcal{Q}_\mathcal{T}$ includes 15 samples per category, with no overlap between the samples in $\mathcal{S}_\mathcal{T}$ and $\mathcal{Q}_\mathcal{T}$. The support set is used to train the classifier, while the query set is utilized to evaluate the performance of the model.

\subsection{Comparison with Other Methods}
To assess the efficacy of our proposed method, we have conducted extensive experiments on four datasets. Table \ref{table_mini}, \ref{table_tiered}, \ref{table_CIFAR-FS}, and \ref{table_CUB} show the performances of some SOTA FSL methods and ours. In these tables, The backbone $a$-$b$-$c$-$d$ denotes a 4-layer convolutional network with $a$, $b$, $c$, and $d$ filters in each layer. Resnet-$n$ refers to a ResNet network with $n$ layers of filters.

\begin{table}[!t]
\begin{center}
\vspace{-20pt}
{\caption{Experimental results (\%) on miniImageNet. The top results are highlighted in bold, and the method with``\dag" denotes that the result of this method is our implementation. The backbones $a$-$b$-$c$-$d$ denotes a 4-layer convolutional network with $a$, $b$, $c$, and $d$ filters in each layer.}
\label{table_mini}}
\setlength{\tabcolsep}{1.6mm}{
\begin{tabular}{lccc}
\hline
\multirow{2}*{\textbf{Method}} & \multirow{2}*{\textbf{Backbone}} & \multicolumn{2}{c} {\textbf{miniImageNet}} \\
&& 5-way 1-shot & 5-way 5-shot \\
\hline
MAML \cite{MAML} & 32-32-32-32 & 48.70 $\pm$ 1.84 & 63.11 $\pm$ 0.92 \\
ProtoNet \cite{ProtoNet} & 64-64-64-64 & $ 49.42 \pm 0.78 $ & 68.20 $\pm$ 0.66 \\
DeepEMD \cite{DeepEMD} & ResNet-12 & 65.91 $\pm$ 0.82 & 82.41 $\pm$ 0.56 \\
RFS-distill \cite{RFS} & ResNet-12 & 64.82 $\pm$ 0.60 & 82.14 $\pm$ 0.43 \\
AssoAlign \cite{AssoAlign} & ResNet-18 & 59.88 $\pm$ 0.67 & 80.35 $\pm$ 0.73 \\
GIFSL \cite{GIFSL} & ResNet-12 & 65.47 $\pm$ 0.63 & 82.75 $\pm$ 0.42 \\
MELR \cite{MELR} & ResNet-12 & 67.40 $\pm$ 0.43 & 83.40 $\pm$ 0.28\\
IEPT \cite{IEPT} & ResNet-12 & 67.05 $\pm$ 0.44 & 82.90 $\pm$ 0.30 \\
IER \cite{IER} & ResNet-12 & 66.82 $\pm$ 0.80 & 84.35 $\pm$ 0.51 \\
RENet \cite{RENet} & ResNet-12 & 67.60 $\pm$ 0.44 & 82.58 $\pm$ 0.30 \\
PAL \cite{PAL} & ResNet-12 & 69.37 $\pm$ 0.64 & 84.40 $\pm$ 0.44 \\
HandCrafted \cite{HandCrafted} & ResNet-12 & 67.14 $\pm$ 0.76 & 83.11 $\pm$ 0.69 \\
PDA \cite{PDA} & ResNet-12 & 65.75 $\pm$ 0.43 & 83.37 $\pm$ 0.30 \\
SCL-distill \cite{Spatial} & ResNet-12 & 67.40 $\pm$ 0.76 & 83.19 $\pm$ 0.54 \\
HGNN \cite{HGNN} & ResNet-12 & 67.02 $\pm$ 0.20 & 83.00 $\pm$ 0.13 \\
APP2S \cite{APP2S} & ResNet-18 & 64.82 $\pm$ 0.12 & 81.31 $\pm$ 0.22 \\
DGAP \cite{DGAP} & ResNet-12 & 61.35 $\pm$ 0.62 & 78.85 $\pm$ 0.46 \\
ESPT \cite{ESPT} & ResNet-12 & 68.36 $\pm$ 0.19 & 84.11 $\pm$ 0.12 \\
Meta-HP \cite{Meta-HP} & ResNet-12 & 62.49 $\pm$ 0.80 & 77.12 $\pm$ 0.62 \\
SAPENet \cite{SAPENet} & ResNet-12 & 66.41 $\pm$ 0.20 & 82.76 $\pm$ 0.14 \\
FEAT+DFR \cite{DFR} & ResNet-12 & 67.74 $\pm$ 0.86 & 82.49 $\pm$ 0.57 \\
MIFN \cite{MIFN} & ResNet-12 & 66.43 $\pm$ 0.63 & 81.51 $\pm$ 0.42 \\
MetaDiff \cite{MetaDiff} & ResNet-12 & 64.99 $\pm$ 0.77 & 81.21 $\pm$ 0.56 \\
IbM2+Meta-Baseline \cite{IbM2} & ResNet-12 & 63.00 $\pm$ 0.00 & 79.50 $\pm$ 0.00 \\
\hline
FEAT \cite{FEAT} & ResNet-12 & 66.78 $\pm$ 0.20 & 82.05 $\pm$ 0.14 \\
\textbf{Ours + FEAT} & ResNet-12 & 69.27 $\pm$ 0.21 & 83.59 $\pm$ 0.13 \\
\hline
Meta-Baseline\dag \cite{MetaBaseline} & ResNet-12 & 63.38 $\pm$ 0.23 & 79.48 $\pm$ 0.16 \\
\textbf{Ours + Meta-Baseline} & ResNet-12 & 69.01 $\pm$ 0.23 & 83.94 $\pm$ 0.15 \\
\hline
STVAE \cite{STVAE} & ResNet-12 & 63.62 $\pm$ 0.80 & 80.68 $\pm$ 0.48 \\
\textbf{Ours + STVAE} & ResNet-12 & 67.29 $\pm$ 0.89 & 82.62 $\pm$ 0.58 \\
\hline
\textbf{Ours} & ResNet-12 & \textbf{69.57 $\pm$ 0.45} & \textbf{84.41 $\pm$ 0.30} \\
\hline
\vspace{-25pt}
\end{tabular}
}
\end{center}
\end{table}

\begin{table}[!t]
\begin{center}
{\caption{Experimental results (\%) on tieredImageNet. The top results are highlighted in bold, and the method with``\dag" denotes that the result of this method is our implementation. The backbones $a$-$b$-$c$-$d$ denotes a 4-layer convolutional network with $a$, $b$, $c$, and $d$ filters in each layer.}
\label{table_tiered}}
\setlength{\tabcolsep}{1.6mm}{
\begin{tabular}{lccc}
\hline
\multirow{2}*{\textbf{Method}} & \multirow{2}*{\textbf{Backbone}} & \multicolumn{2}{c} {\textbf{tieredImageNet}} \\
& & 5-way 1-shot & 5-way 5-shot \\
\hline
MAML \cite{MAML} & 32-32-32-32 & 51.67 $\pm$ 1.81 & 70.30 $\pm$ 1.75 \\
ProtoNet \cite{ProtoNet} & 64-64-64-64  & 53.31 $\pm$ 0.89 & 72.69 $\pm$ 0.74 \\
DeepEMD \cite{DeepEMD} & ResNet-12 & 71.16 $\pm$ 0.87 & 86.03 $\pm$ 0.58 \\
RFS-distill \cite{RFS} & ResNet-12 & 71.52 $\pm$ 0.69 & 86.03 $\pm$ 0.49 \\
AssoAlign \cite{AssoAlign} & ResNet-18 & 69.29 $\pm$ 0.56 & 85.97 $\pm$ 0.49 \\
GIFSL \cite{GIFSL} & ResNet-12 & 72.39 $\pm$ 0.66 & 86.91 $\pm$ 0.44 \\
MELR \cite{MELR} & ResNet-12 & 72.14 $\pm$ 0.51 & 87.01 $\pm$ 0.35 \\
IEPT \cite{IEPT} & ResNet-12 & 72.24 $\pm$ 0.50 & 86.73 $\pm$ 0.34 \\
IER \cite{IER} & ResNet-12 & 71.87 $\pm$ 0.89 & 86.82 $\pm$ 0.58 \\
RENet \cite{RENet} & ResNet-12 & 71.16 $\pm$ 0.51 & 85.28 $\pm$ 0.35 \\
PAL \cite{PAL} & ResNet-12 & 72.25 $\pm$ 0.72 & 86.95 $\pm$ 0.47 \\
PDA \cite{PDA} & ResNet-12 & 72.28 $\pm$ 0.49 & 86.70 $\pm$ 0.33 \\
SCL-distill \cite{Spatial} & ResNet-12 & 71.98 $\pm$ 0.91 & 86.19 $\pm$ 0.59 \\
HGNN \cite{HGNN} & ResNet-12 & 72.05 $\pm$ 0.23 & 86.49 $\pm$ 0.15 \\
APP2S \cite{APP2S} & ResNet-18 & 70.83 $\pm$ 0.15 & 84.15 $\pm$ 0.29 \\
DGAP \cite{DGAP} & ResNet-12 & 70.10 $\pm$ 0.67 & 84.99 $\pm$ 0.46 \\
ESPT \cite{ESPT} & ResNet-12 & 72.68 $\pm$ 0.22 & \textbf{87.49 $\pm$ 0.14} \\
Meta-HP \cite{Meta-HP} & ResNet-12 & 68.26 $\pm$ 0.72 & 82.91 $\pm$ 0.36 \\
SAPENet \cite{SAPENet} & ResNet-12 & 68.63 $\pm$ 0.23 & 84.30 $\pm$ 0.16 \\
FEAT+DFR \cite{DFR} & ResNet-12 & 71.31 $\pm$ 0.93 & 85.12 $\pm$ 0.64 \\
MIFN \cite{MIFN} & ResNet-12 & 70.03 $\pm$ 0.72 & 84.14 $\pm$ 0.50 \\
MetaDiff \cite{MetaDiff} & ResNet-12 & 72.33 $\pm$ 0.92 & 86.31 $\pm$ 0.62 \\
\hline
FEAT \cite{FEAT} & ResNet-12 & 70.80 $\pm$ 0.23 & 84.79 $\pm$ 0.16 \\
\textbf{Ours + FEAT} & ResNet-12 & 72.02 $\pm$ 0.23  & 86.19 $\pm$ 0.15 \\
\hline
Meta-Baseline\dag \cite{MetaBaseline} & ResNet-12 & 68.74 $\pm$ 0.26 & 83.45 $\pm$ 0.18 \\
\textbf{Ours + Meta-Baseline} & ResNet-12 & 69.79 $\pm$ 0.26 & 83.55 $\pm$ 0.18 \\
\hline
STVAE\dag \cite{STVAE} & ResNet-12 & 68.32 $\pm$ 0.94 & 83.79 $\pm$ 0.66 \\
\textbf{Ours + STVAE} & ResNet-12 & 72.03 $\pm$ 0.89 & 84.49 $\pm$ 0.66 \\
\hline
\textbf{Ours} & ResNet-12 & \textbf{72.98 $\pm$ 0.51} & 86.23 $\pm$ 0.34 \\
\hline
\vspace{-25pt}
\end{tabular}
}
\end{center}
\end{table}

\begin{table}[!h]
\begin{center}
\vspace{5pt}
{\caption{Experimental results (\%) on CIFAR-FS. The top results are highlighted in bold, and the method with``\dag" denotes that the result of this method is our implementation. The backbones $a$-$b$-$c$-$d$ denotes a 4-layer convolutional network with $a$, $b$, $c$, and $d$ filters in each layer.}
\label{table_CIFAR-FS}}
\setlength{\tabcolsep}{1.6mm}{
\begin{tabular}{lccc}
\hline
\textbf{Method} & \textbf{Backbone} & 5-way 1-shot & 5-way 5-shot \\
\hline
MAML \cite{MAML} & 32-32-32-32 & 58.90 $\pm$ 1.90 & 71.50 $\pm$ 1.00 \\
ProtoNet \cite{ProtoNet} & 64-64-64-64 & 55.50 $\pm$ 0.70 & 72.00 $\pm$ 0.60 \\
RFS-distill \cite{RFS} & ResNet-12 & 73.90 $\pm$ 0.80 & 86.90 $\pm$ 0.50  \\
GIFSL \cite{GIFSL} & ResNet-12 & 74.58 $\pm$ 0.38 & 87.68 $\pm$ 0.23 \\
IER \cite{IER} & ResNet-12 & 76.83 $\pm$ 0.82 & 89.26 $\pm$ 0.58  \\
RENet \cite{RENet} & ResNet-12 & 74.51 $\pm$ 0.46 & 86.60 $\pm$ 0.32  \\
PAL \cite{PAL} & ResNet-12 & 77.10 $\pm$ 0.70 & 88.00 $\pm$ 0.50  \\
HandCrafted \cite{HandCrafted} & ResNet-12 & 76.68 $\pm$ 0.59 & 87.49 $\pm$ 0.73 \\
SCL-distill \cite{Spatial} & ResNet-12 & 76.50 $\pm$ 0.90 & 88.00 $\pm$ 0.60  \\
ConstellationNet \cite{ConstellationNet} & ResNet-12 & 75.40 $\pm$ 0.20 & 86.80 $\pm$ 0.20  \\
APP2S \cite{APP2S} & ResNet-18 & 73.12 $\pm$ 0.22 & 85.69 $\pm$ 0.16 \\
Meta-HP \cite{Meta-HP} & ResNet-12 & 73.74 $\pm$ 0.57 & 86.37 $\pm$ 0.32 \\
IbM2+Meta-Baseline \cite{IbM2} & ResNet-12 & 72.30 $\pm$ 0.00 & 85.10 $\pm$ 0.00 \\
\hline
FEAT\dag \cite{FEAT} & ResNet-12 & 75.97 $\pm$ 0.21 & 87.34 $\pm$ 0.14 \\
\textbf{Ours + FEAT} & ResNet-12 & 79.91 $\pm$ 0.21 & \textbf{90.18 $\pm$ 0.14} \\
\hline
Meta-Baseline\dag \cite{MetaBaseline} & ResNet-12 & 74.56 $\pm$ 0.39 & 86.24 $\pm$ 0.27 \\
\textbf{Ours + Meta-Baseline} & ResNet-12 & 78.51 $\pm$ 0.24 & 88.60 $\pm$ 0.16 \\
\hline
STVAE \cite{STVAE} & ResNet-12 & 76.30 $\pm$ 0.60 & 87.00 $\pm$ 0.40 \\
\textbf{Ours + STVAE} & ResNet-12 & \textbf{80.92 $\pm$ 0.72} & 86.38 $\pm$ 0.60 \\
\hline
\textbf{Ours} & ResNet-12 & 78.54 $\pm$ 0.47 & 88.64 $\pm$ 0.32 \\
\hline
\vspace{-25pt}
\end{tabular}
}
\end{center}
\end{table}

\vspace{10pt}
\noindent\textbf{General Few-Shot Classification.} 
As shown in Table \ref{table_mini} and \ref{table_tiered}, our method achieves promising performance compared to other methods on miniImageNet and tieredImageNet. To be specific, our method achieves 69.57$\%$, 84.41$\%$ for 1-shot and 5-shot tasks on miniImageNet, and 72.98$\%$, 86.23$\%$ on tieredImageNet. Especially in 5-way 1-shot FSL tasks, our method achieves SOTA. On CIFAR-FS, our method achieves 78.54$\%$, 88.64$\%$ in 1-shot and 5-shot FSL tasks respectively, as shown in Table \ref{table_CIFAR-FS}. Note that, unlike RFS \cite{RFS}, PAL \cite{PAL}, and SCL \cite{Spatial}, which adopt the knowledge distillation technique, and DeepEMD \cite{DeepEMD}, ESPT \cite{ESPT}, which use meta-learning method, our method does not need the second training or the meta-tuning phase. Our approach enables a pre-trained model to achieve performance that is comparable to and even exceeds, that of SOTA methods.

\vspace{10pt}
\noindent\textbf{Fine-Grained Few-Shot Classification.}
Moreover, to further validate the generalizability of our approach, we also conduct experiments on a fine-grained dataset, CUB. The experimental results demonstrate that our approach outperforms all other methods, as shown in Table \ref{table_CUB}. In particular, our method achieves 86.14$\%$ in 1-shot tasks and 94.75$\%$ in 5-shot tasks respectively, outperforming the second-best results of 0.69$\%$ and 0.73$\%$. These results indicate that on fine-grained datasets with small category differences, our approach can better distinguish fine-grained categories by exploring sample relations at different granularities and modeling them in detail.

\begin{table}[!h]
\begin{center}
{\caption{Experimental results (\%) on CUB. The top results are highlighted in bold, and the method with``\dag" denotes that the result of this method is our implementation.The backbones $a$-$b$-$c$-$d$ denotes a 4-layer convolutional network with $a$, $b$, $c$, and $d$ filters in each layer.}
\label{table_CUB}}
\setlength{\tabcolsep}{2.0mm}{
\begin{tabular}{lccc}
\hline
\textbf{Method} & \textbf{Backbone} & 5-way 1-shot & 5-way 5-shot \\
\hline
FEAT \cite{FEAT} & 64-64-64-64 & 68.87 $\pm$ 0.22 & 82.90 $\pm$ 0.15 \\
DeepEMD \cite{DeepEMD} & ResNet-12 & 75.65 $\pm$ 0.83 & 88.69 $\pm$ 0.50 \\
AssoAlign \cite{AssoAlign} & ResNet-18 & 74.22 $\pm$ 1.09 & 88.65 $\pm$ 0.55 \\
MELR \cite{MELR} & 64-64-64-64 & 70.26 $\pm$ 0.50 & 85.01 $\pm$ 0.32 \\
IEPT \cite{IEPT} & 64-64-64-64 & 69.97 $\pm$ 0.49 & 84.33 $\pm$ 0.33 \\
RENet \cite{RENet} & ResNet-12 & 79.49 $\pm$ 0.44 & 91.11 $\pm$ 0.24 \\
HGNN \cite{HGNN} & ResNet-12 & 78.58 $\pm$ 0.20 & 90.02 $\pm$ 0.12 \\
APP2S \cite{APP2S} & ResNet-12 & 77.64 $\pm$ 0.19 & 90.43 $\pm$ 0.18 \\
ESPT \cite{ESPT} & ResNet-12 & 85.45 $\pm$ 0.18 & 94.02 $\pm$ 0.09 \\
SAPENet \cite{SAPENet} & 64-64-64-64 & 70.38 $\pm$ 0.23 & 84.47 $\pm$ 0.14 \\
FEAT+DFR \cite{DFR} & ResNet-12 & 77.14 $\pm$ 0.21 & 88.97 $\pm$ 0.13 \\
\hline
FEAT\dag \cite{FEAT} & ResNet-12 & 77.60 $\pm$ 0.45 & 89.20 $\pm$ 0.28 \\
\textbf{Ours + FEAT} & ResNet-12 & 84.23 $\pm$ 0.19 & 92.67 $\pm$ 0.10 \\
\hline
Meta-Baseline\dag \cite{MetaBaseline} & ResNet-12 & 75.04 $\pm$ 0.24 & 87.57 $\pm$ 0.14 \\
\textbf{Ours + Meta-Baseline} & ResNet-12 & \textbf{88.37 $\pm$ 0.18} & \textbf{95.52 $\pm$ 0.09} \\
\hline
STVAE \cite{STVAE} & ResNet-12 & 77.32 $\pm$ 0.00 & 86.84 $\pm$ 0.00\\
\textbf{Ours + STVAE} & ResNet-12 & 84.35 $\pm$ 0.76 & 93.69 $\pm$ 0.39 \\
\hline
\textbf{Ours} & ResNet-12 & 86.14 $\pm$ 0.38 & 94.75 $\pm$ 0.19 \\
\hline
\vspace{-20pt}
\end{tabular}
}
\end{center}
\end{table}

\vspace{10pt}
\noindent\textbf{Combination with Other Methods.} \label{combination}
Additionally, as a feature-learning-based approach, our work can provide a good pre-trained model for two-stage meta-learning methods and some generative methods, helping them to achieve better performance. To demonstrate this, we select two meta-learning methods (FEAT \cite{FEAT}, Meta-Baseline \cite{MetaBaseline}) and a generative method (STVAE \cite{STVAE}) to conduct experiments on four datasets. As shown in Table \ref{table_mini} and \ref{table_tiered}, when using our pre-trained model, FEAT, Meta-Baseline, and STVAE achieve improvements of 2.49\%, 5.63\%, and 3.67\% respectively in 1-shot tasks on miniImageNet, and improvements of 1.54\%, 4.46\%, 1.94\% in 5-shot tasks. FEAT, Meta-Baseline, and STVAE also have improvements on tieredImageNet using our model as their pre-trained model. On CIFAR-FS, STVAE and FEAT using our pre-trained model achieve the best result in 1-shot and 5-shot respectively, which are 80.92\% and 90.18\%, as shown in Table \ref{table_CIFAR-FS}. Furthermore, we also conduct experiments for these approaches on CUB, where 
Meta-Baseline stands out by delivering the highest performance in both 1-shot and 5-shot FSL tasks, achieving accuracies of 88.37\% and 95.52\% respectively, as shown in Table \ref{table_CUB}. In these experiments, results with ``\dag" denote our re-implementation of the method, because the backbones of these approaches have some differences with ours or they have not conducted experiments on the corresponding dataset. These experimental results indicate that our approach can provide a good pre-trained model for these two-stage meta-learning methods and generative methods to improve their performance.

\begin{table}[!h]
\begin{center}
{\caption{Few-shot classification accuracy (in \%) on miniImageNet and CUB with different components.}
\label{table_ablation_module}}
\setlength{\tabcolsep}{4.3mm}{
\begin{tabular}{lcc}
\hline
\multirow{2}*{\textbf{Method}} & \multicolumn{2}{c}{\textbf{miniImageNet}} \\
& 5-way 1-shot & 5-way 5-shot \\
\hline
Baseline & 66.78 $\pm$ 0.43 & 83.82 $\pm$ 0.29 \\
Baseline w/ SS &  67.76 $\pm$ 0.44 & 84.31 $\pm$ 0.28 \\
Baseline w/ TCL & 68.45 $\pm$ 0.44 & 84.37 $\pm$ 0.29 \\
Baseline w/ CCL & 68.61 $\pm$ 0.44 & 84.13 $\pm$ 0.29 \\
Baseline w/ TCL \& CCL & 69.21 $\pm$ 0.45 & 84.37 $\pm$ 0.30 \\
Baseline w/ all & 69.57 $\pm$ 0.45 & 84.41 $\pm$ 0.30 \\
\hline
\hline
\multirow{2}*{\textbf{Method}} & \multicolumn{2}{c}{\textbf{CUB}}\\
& 5-way 1-shot & 5-way 5-shot \\
\hline
Baseline & 82.18 $\pm$ 0.40 & 93.70 $\pm$ 0.20 \\
Baseline w/ SS & 83.46 $\pm$ 0.39 & 94.18 $\pm$ 0.20 \\
Baseline w/ TCL & 83.16 $\pm$ 0.39 & 93.74 $\pm$ 0.20 \\
Baseline w/ CCL & 85.30 $\pm$ 0.38 & 94.50 $\pm$ 0.19 \\
Baseline w/ TCL \& CCL & 85.53 $\pm$ 0.38 & 94.30 $\pm$ 0.19 \\
Baseline w/ all & 86.14 $\pm$ 0.38 & 94.75 $\pm$ 0.19 \\
\hline
\vspace{-20pt}
\end{tabular}
}
\end{center}
\end{table}

\subsection{Component Ablative Analysis}
In order to study the impact of each component, we conduct comprehensive ablation studies on miniImageNet and CUB. Here, our baseline model is the same as RFS \cite{RFS}, but we add some augmented samples in order to alleviate the overfitting problems and implement our Transformation Consistency Learning (TCL).

As shown in Table \ref{table_ablation_module}, our baseline achieves 66.78\% and 82.18\% in 5-way 1-shot FSL tasks on miniImageNet and CUB, respectively. When adding the self-supervised component to predict which image transformation was performed, we obtain improvements of 0.98\% and 1.28\% over the baseline on miniImageNet and CUB respectively. By enforcing the intra-sample relation of the same sample under different transformations (adding TCL component on the baseline), we obtain improvements of 1.67\% and 0.98\% on miniImageNet and CUB respectively. By enforcing the intra-class relation of homogenous samples and the inter-class relation of inhomogeneous samples (adding CCL component on the baseline), we obtain improvements of 1.83\% and 3.12\% over the baseline on miniImageNet and CUB respectively. Besides, for 5-way 5-shot FSL tasks, adding different components also leads to results that are either superior or comparable to those of the baseline. On the CUB, the results of adding CCL are better than adding the other components, due to the fact that CUB is a fine-grained dataset, in which the differences between different classes are relatively small, and pushing them far away works better on CUB compared to on miniImageNet. 

Additionally, by using both TCL and CCL components, our model achieves better performance than only using one in 5-way 1-shot tasks, the accuracies achieve 69.21\% and 85.53\% on miniImageNet and CUB, respectively. And when using all three components (TCL, CCL, and SS), our model achieves the best performance on miniImageNet and CUB, which are 69.57\%, 86.14\% in 1-shot FSL tasks, and 84.41\%, 94.75\% in 5-shot FSL tasks. Overall, these experimental results demonstrate the effectiveness of each component of our approach.

\subsection{Parameter Ablative Studies}
\noindent\textbf{Effects of hyper-parameters $\alpha$ and $\beta$.} $\alpha$ and $\beta$ are the hyper-parameters used to adjust the weights of different losses. Here, we evaluate the model performance on miniImageNet by assigning various values to $\alpha$ and $\beta$. When discussing the effect of one hyper-parameter, we set the other parameter to 0.

\begin{figure}[h]
  \centering
  \begin{subfigure}{0.49\textwidth}
    \includegraphics[width=\textwidth]{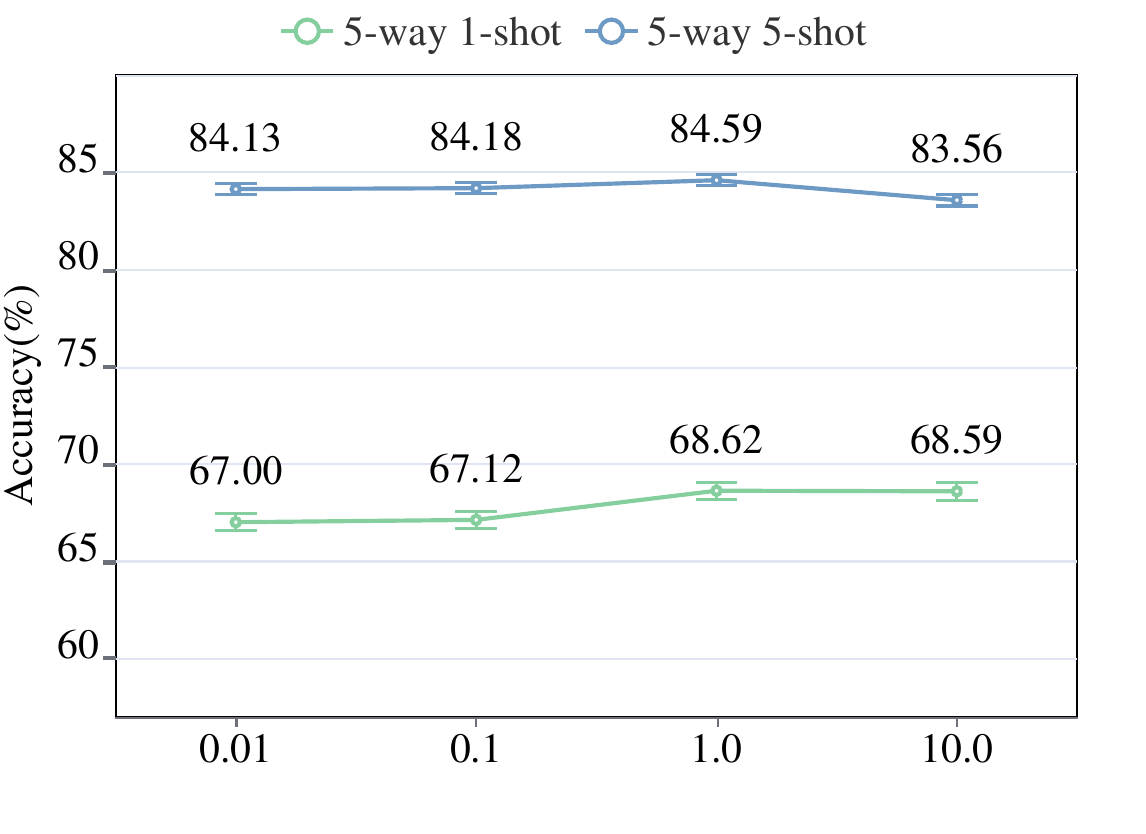}
    \vspace{-25pt}
    \caption{$\alpha$}
    \label{fig_alpha}
  \end{subfigure}
  \hfill
  \begin{subfigure}{0.49\textwidth}
    \includegraphics[width=\textwidth]{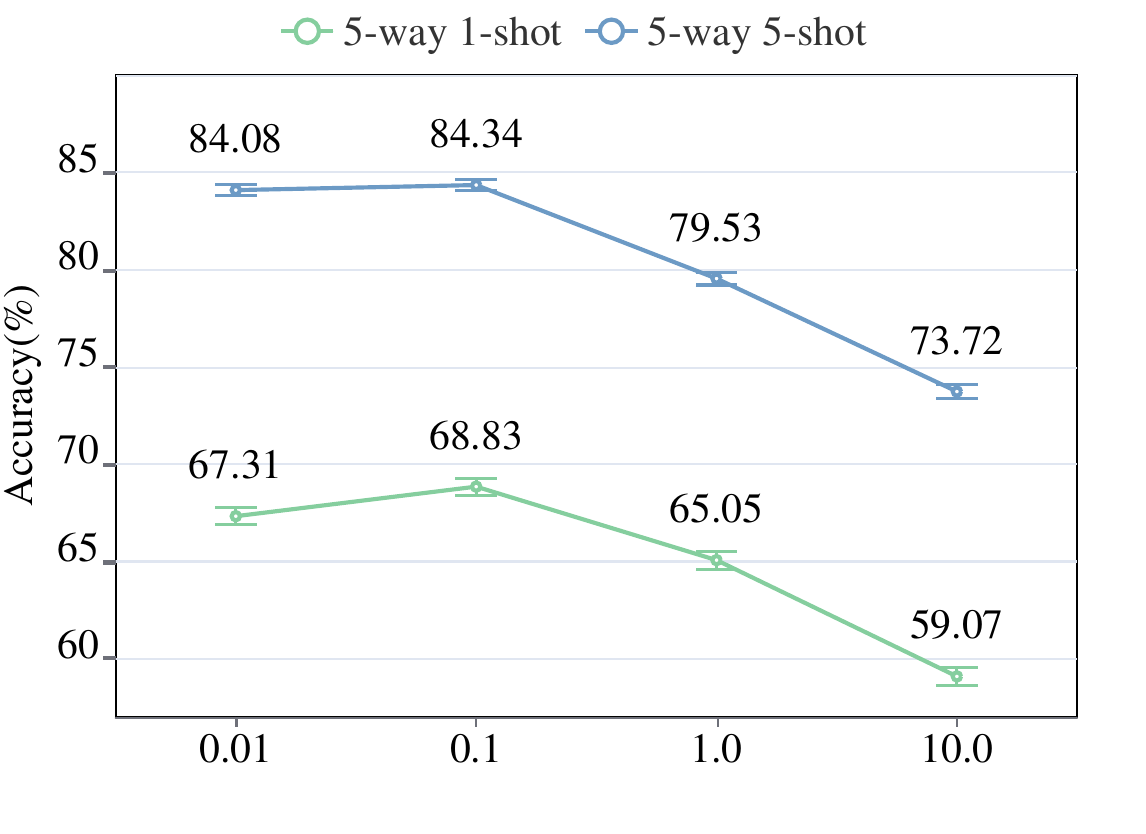}
    \vspace{-25pt}
    \caption{$\beta$}
    \label{fig_beta}
  \end{subfigure}
  \caption{Effects of hyper-parameters $\alpha$ and $\beta$ on miniImageNet. (a) $\alpha$. (b) $\beta$.}
  \label{fig_alpha_beta}
\end{figure}

\begin{table}[h]
    \centering
    \caption{Effects of hyper-parameters $\alpha$ and $\beta$ on miniImageNet.}
    \label{tab:alpha_beta}
    \begin{tabular}{ccccc}
        \toprule
        \diagbox{${\alpha}$}{${\beta}$} & 0.01 & 0.1 & 1.0 & 10.0 \\
        \midrule
        0.01 & 67.57 $\pm$ 0.43 & 68.76 $\pm$ 0.44 & 65.71 $\pm$ 0.46 & 58.77 $\pm$ 0.47 \\
        0.1 & 67.07 $\pm$ 0.43 & 68.82 $\pm$ 0.44 & 65.88 $\pm$ 0.46 & 59.53 $\pm$ 0.48 \\
        1.0 & 68.95 $\pm$ 0.44 & \textbf{69.57 $\pm$ 0.45} & 66.68 $\pm$ 0.48 & 59.40 $\pm$ 0.48 \\
        10.0 & 68.37 $\pm$ 0.46 & 69.22 $\pm$ 0.47 & 67.16 $\pm$ 0.48 & 59.25 $\pm$ 0.48 \\
        \bottomrule
    \end{tabular}
\end{table}

As shown in Fig. \ref{fig_alpha_beta}, for the hyper-parameter $\alpha$, the change in model performance is not significant. The best performance is achieved when $\alpha = 1.0$. And for the hyper-parameter $\beta$, we observe that the performance of the model initially improves and then deteriorates with an increase in $\beta$. Moreover, we employ grid search to determine the optimal value of the hyperparameter ${\alpha}$ and ${\beta}$. We have conducted experiments to discuss the optimal value of them, and experimental results demonstrate that the model reaches the best performance when ${\alpha}$ and ${\beta}$ is set to 1.0 and 0.1 respectively, As shown in Table \ref{tab:alpha_beta}. Therefore, we set $\alpha = 1.0$, and $\beta = 0.1$ in our final model.

\begin{figure}[h]
  \centering
  \begin{subfigure}{0.49\textwidth}
    \includegraphics[width=\textwidth]{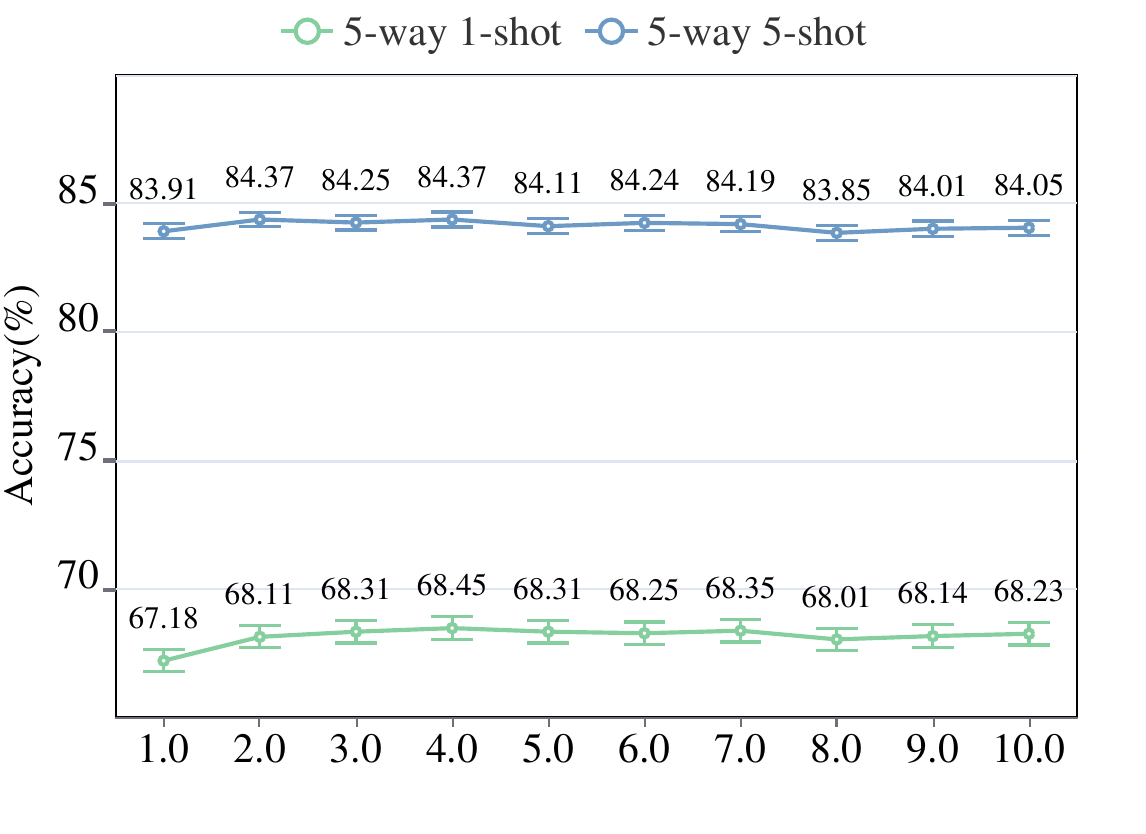}
    \vspace{-25pt}
    \caption{$\tau_1$}
    \label{fig_τ1}
  \end{subfigure}
  \hfill
  \begin{subfigure}{0.49\textwidth}
    \includegraphics[width=\textwidth]{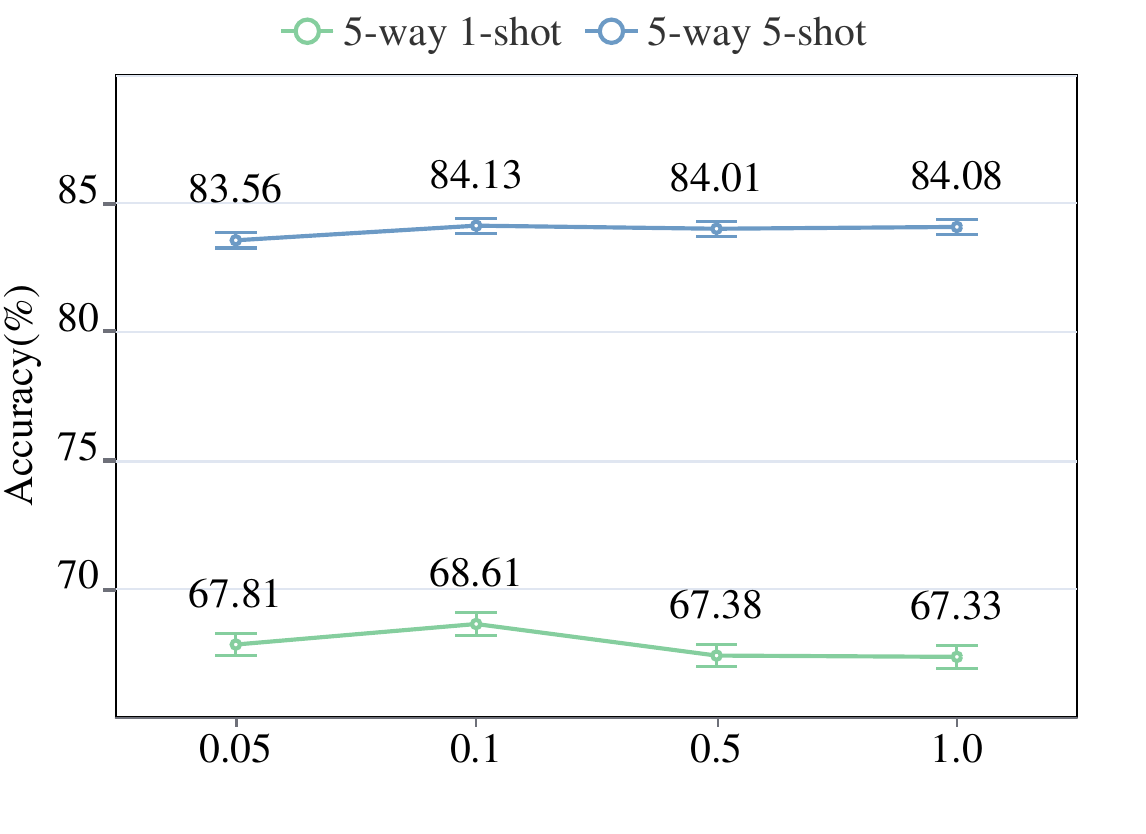}
    \vspace{-25pt}
    \caption{$\tau_2$}
    \label{fig_τ2}
  \end{subfigure}
  \caption{Effects of hyper-parameters $\tau_1$ and $\tau_2$ on miniImageNet. (a) $\tau_1$. (b) $\tau_2$.}
  \label{fig_tau}
\end{figure}

\vspace{10pt}
\noindent\textbf{Effects of Temperature Parameters $\tau_1$ and $\tau_2$.} Additionally, we also discuss the impact of the temperature parameters $\tau_1$ and $\tau_2$ on the experimental results. Firstly, $\tau_1$ is used to smooth the predicted output to provide more information. We set it from 1.0 to 10.0 to evaluate our model on miniImageNet. As shown in Fig. \ref{fig_τ1}, when $\tau_1$ varies from 1.0 to 10.0, the experimental results do not change significantly. When $\tau_1$ is set to 4.0, the optimal result is achieved. $\tau_2$ is the temperature parameter used in CCL component. We evaluate the model performance when $\tau_2$ is set to 0.05, 0.1, 0.5, and 1.0. As shown in Fig. \ref{fig_τ2}, the optimal result is achieved when $\tau_2 = 0.1$. So, we set $\tau_1 = 4.0$, and $\tau_2 = 0.1$ in our final model.


\begin{figure}[h]
  \centering
  \begin{subfigure}{0.49\textwidth}
    \includegraphics[width=\textwidth]{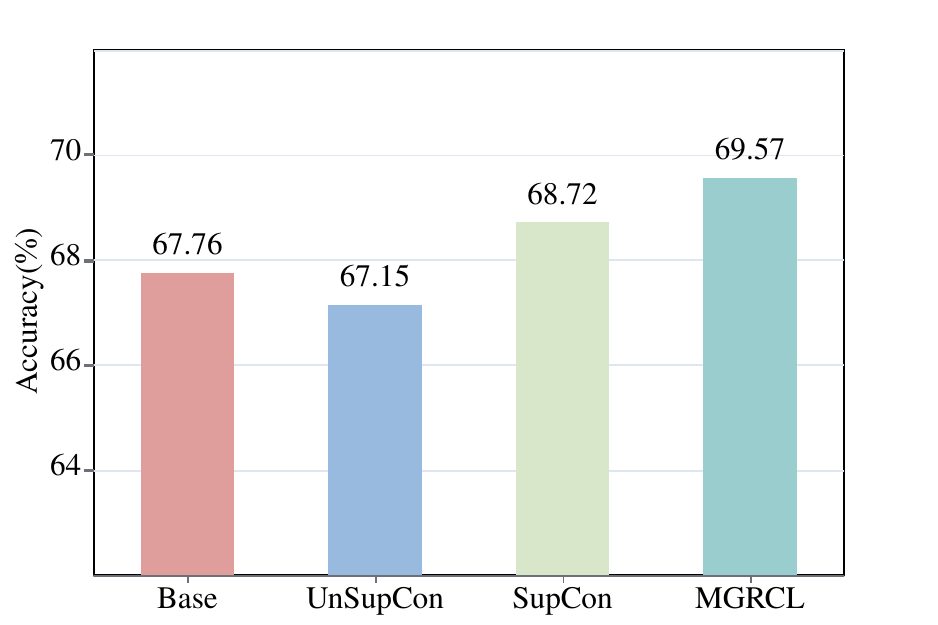}
    \caption{miniImageNet}
    \label{fig_mnImageNet}
  \end{subfigure}
  \hfill
  \begin{subfigure}{0.49\textwidth}
    \includegraphics[width=\textwidth]{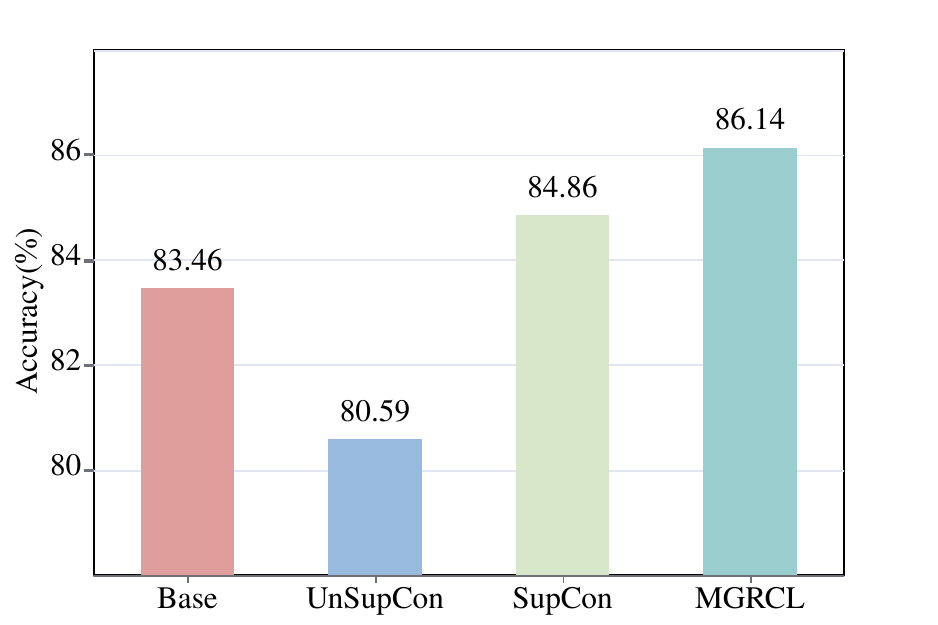}
    \caption{CUB}
    \label{fig_cub}
  \end{subfigure}
  \caption{Comparison to the base learner with Unsupervised Contrastive Learning or Supervised Contrastive Learning in 5-way 1-shot tasks on miniImageNet and CUB. (a) miniImageNet. (b) CUB.}
  \label{fig_comparison}
\end{figure}

\subsection{Sample Relation Exploitation Strategy Discussion}
In recent years, some FSL approaches use contrastive learning to exploit sample relations. However, these methods often directly use unsupervised learning (UnSupCon) \cite{SimCLR} or supervised contrastive learning (SupCon) \cite{SupCon}, which is not appropriate because they do not fully exploit the sample relations. To indicate the superiority of our method in comparison to these methods, we conduct experiments based on the same base learner we proposed.\footnote{Here we use the code provided by SupCon to implement the unsupervised contrastive learning method (SimCLR), and the supervised contrastive learning method (SupCon). The source code at \href{https://github.com/HobbitLong/SupContrast}{https://github.com/HobbitLong/SupContrast}.} During implementation, the UnSupCon or SupCon loss is added directly to the original loss as an auxiliary loss like our method. As shown in Fig. \ref{fig_comparison}, the performance adding UnSupCon compared to the Base even decreased on both datasets. This can be attributed to the fact that UnSupCon treats the transformations of the anchor image as positive samples, and other images as negative samples. As a result, it pushes the homogenous samples apart, leading to a decrease in performance. On the other hand, incorporating SupCon into the model does not suffer from this issue. However, SupCon treats the transformations of one sample and its homogenous samples as the same relation, which is inappropriate because the different transformations of one sample should have the same semantic content, while the semantic content of homogenous samples should be only similar but not exactly consistent. In contrast, our proposed approach fully considers the sample relations at different granularities and models them in detail, thus achieving the best performance on the same base learner. This indicates the effectiveness of our approach in leveraging sample relations.

\begin{figure}[h]
  \centering
  \begin{subfigure}{0.24\textwidth}
    \includegraphics[width=\textwidth]{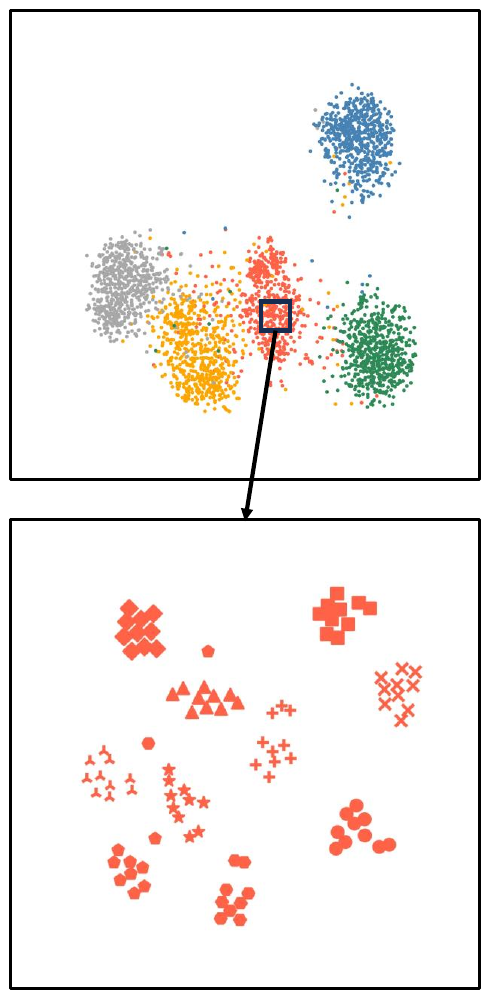}
    \caption{Base}
    \label{fig_1}
  \end{subfigure}
  \begin{subfigure}{0.24\textwidth}
    \includegraphics[width=\textwidth]{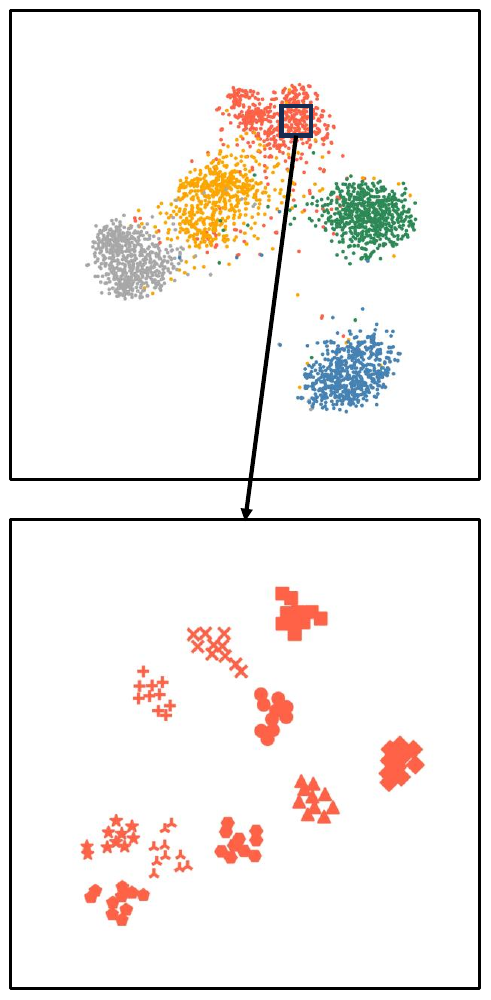}
    \caption{Base+TCL}
    \label{fig_2}
  \end{subfigure}
  \begin{subfigure}{0.24\textwidth}
    \includegraphics[width=\textwidth]{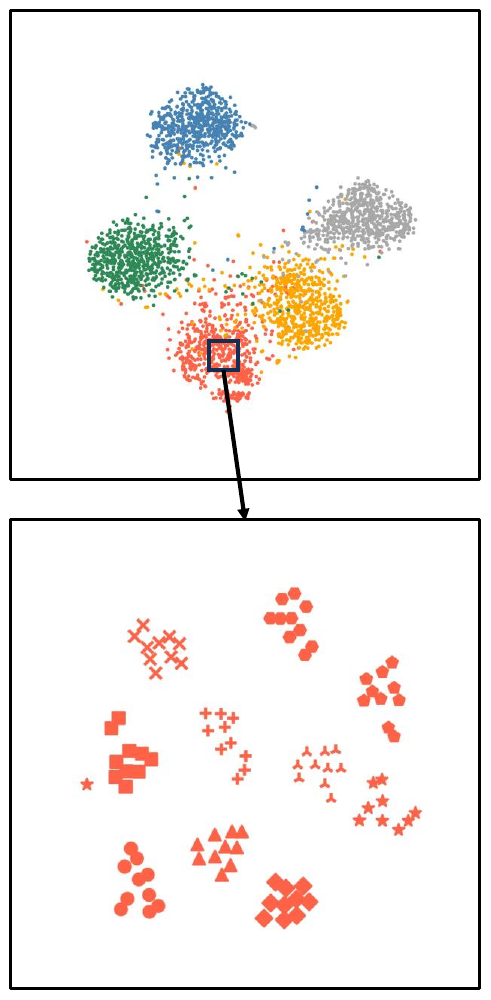}
    \caption{Base+CCL}
    \label{fig_3}
  \end{subfigure}
  \begin{subfigure}{0.24\textwidth}
    \includegraphics[width=\textwidth]{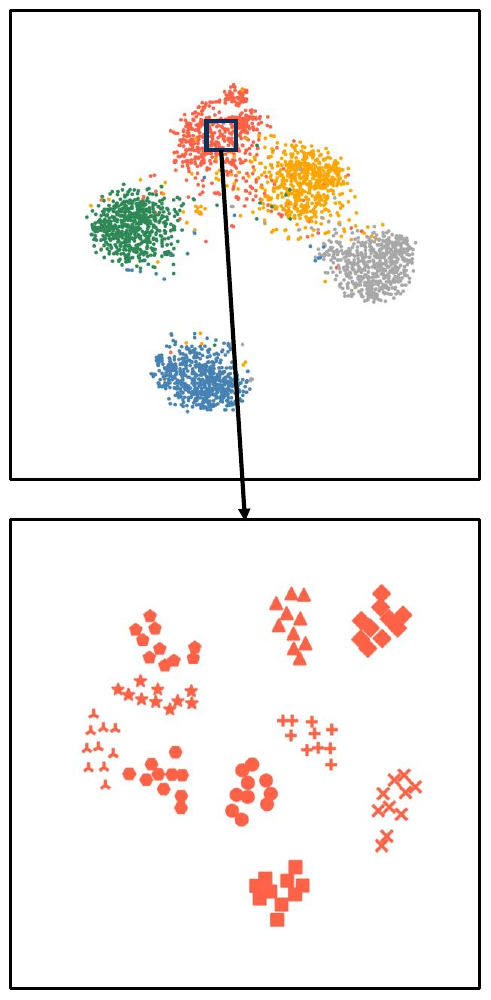}
    \caption{Ours}
    \label{fig_4}
  \end{subfigure}
  \caption{t-SNE visualization results for 5 randomly selected novel classes of miniImageNet. Here, different colors and shapes represent different classes and samples, respectively. The first row shows the intra-class and inter-class relations, and the second row shows the intra-sample realtion. In order to more conveniently check the quality of the classification boundary in the first row, we give numerical information: $d_1/d_2$, where $d_1$ represents the mean of the average distance between the samples and the sample center of 5 classes (the degree of sample cohesion within the class), $d_2$ represents the average of the distance between each class center (the degree of divergence of different class centers). (a) Base: 1.01, (b) Base + TCL: 0.99, (c) Base + CCL: 0.91, (d) Ours: 0.90.}
  \label{fig_tsne}
\end{figure}

\subsection{Visualization Analysis}
To better demonstrate the effectiveness of our model, we conduct visualization experiments using t-SNE on miniImageNet, as shown in Figure \ref{fig_tsne}. In this figure, we can observe that the Base model already has a definite classification boundary. And compared with the Base model and Base with TCL, the samples within the class exhibit better cohesion in the model of Base with CCL and the final model, which prove the effevtiveness of enforcing the intra-class relation of homogenous samples and the inter-class relation of inhomogeneous samples. In addition, by enforcing the intra-sample relation of the same sample under different transformation, the features of the different transformations retain consistent, like in the second of Figure \ref{fig_2} and Figure \ref{fig_4}. While in the other models, there are features of some transformations are far away from the features of other transformations, which demonstrates the role of TCL component.


\section{Discussions}
\subsection{Limitions in Real Scenarios}
Real-world scenarios, such as industrial defect detection, medical image classification, and clinical settings, often involve limited labeled samples and present challenges distinct from those encountered in academic datasets. Discrepancies between real-world data and few-shot datasets, such as insufficient base class data for pre-training, make the applications of few-shot classification in real-world scenarios more difficult. Additionally, real-world applications require higher accuracy and reliability, and the lack of labeled data exacerbates these issues. Domain shifts, dataset biases, and the need for robust models are often overlooked in current few-shot learning research. Moving forward, we aim to apply our proposed method and other few-shot algorithms to real-world scenarios to gain deeper insights into their performance and limitations in practical applications.

\subsection{Computational Resources}
First, the method we proposed makes only minimal changes to the network architecture, primarily by adding a multilayer perceptron. Therefore, the network structure does not introduce significant computational overhead. Additionally, while the method feeds different data-augmented samples into the network simultaneously, which does incur extra computational cost, this approach is commonly used in most current few-shot classification methods to train a feature extraction network with strong transferability on base datasets such as \cite{IER, PAL, ESPT, FEAT}, etc. This additional cost is considered acceptable, as the current focus in few-shot learning (FSL) lies more on improving accuracy than minimizing computational resource consumption. To further reduce computational costs, we plan to explore more advanced data augmentation techniques, such as Mixup and Mosaic, which enable the model to acquire richer knowledge from fewer samples.

Moreover, the memory bank used in our approach does indeed depend on the size of the dataset, as it stores feature vectors for all images. One solution is to use a Momentum Encoder, similar to MoCo \cite{MoCo}, which would eliminate the need to store feature vectors for each image. However, this approach introduces additional computational cost, as each batch would require a forward propagation through the Momentum Encoder to obtain image features. In this paper, we choose to prioritize computational cost over memory consumption. Another solution we are currently exploring is to store only one prototype feature per class and add noise to the prototype features to simulate sample variability, thus mitigating excessive memory usage. 

To reduce computational resource consumption and more elegantly apply the proposed method to larger datasets, we will continue to explore more suitable data augmentation methods and engineering implementations in future work to make this approach more effectively scalable to larger real-world datasets.

\subsection{Comparison with Existing Models}
Compared to other methods, the proposed approach is a simple and effective pre-training method that fully leverages the relationships between different samples. It achieves results that are either better than or comparable to other methods with a single training phase \cite{IEPT, IER, HandCrafted}, and reducing the training complexity compared to two-stage methods \cite{FEAT, MetaBaseline, STVAE}. Furthermore, it can serve as a good feature extractor for some two-stage methods, enhancing their performance, as demonstrated in Section \ref{combination}. Moreover, similar to many few-shot classification methods, such as \cite{RFS, IER, HandCrafted, Spatial}, our approach utilizes the feature extraction network obtained during the pre-training process to extract image features for both the support set and the query set during testing phase. A logistic regression classifier is then trained on the support set features to classify the query set features. Since the feature extraction network and the testing procedure are the same, the inference speed of our method is essentially consistent with that of most few-shot classification approaches.

However, our method has its limitations, the most notable being the use of a larger number of data-augmented samples, which increases computational resource consumption. To address this limitation, we plan to explore advanced data augmentation techniques that allow the model to acquire richer knowledge from fewer samples, thereby mitigating the computational overhead.

\section{Conclusion} \label{conclu}
In this paper, we rethink sample relations and categorize them into three distinct types at different granularities: intra-sample relation of the same sample under different transformations, intra-class relation of homogenous samples, and inter-class relation of inhomogeneous samples. By exploiting these relations, we introduce a straightforward yet effective contrastive learning approach, Multi-Grained Relation Contrastive Learning (MGRCL), for boosting few-shot learning. In MGRCL, we design Transformation Consistency Learning (TCL) to directly enforce different transformations of the same sample to have the same semantic content by label distribution alignment, and Class Contrastive Learning (CCL) to indirectly encourage closer feature proximity for homogenous samples while pushing features of inhomogeneous samples further apart. Experimental results on four FSL benchmarks demonstrate the effectiveness of our proposed method. Moreover, our approach can provide a good pre-trained model for these two-stage FSL methods and improve their performance.

\section{Acknowledgements}
The reported research is partly supported by the National Natural Science Foundation of China (No.62176030, 62276033), and the Natural Science Foundation of Chongqing under Grant cstc2021jcyj-msxmX0568.

\bibliographystyle{elsarticle-num}
\bibliography{reference}

\begin{thebibliography}{10}
\expandafter\ifx\csname url\endcsname\relax
  \def\url#1{\texttt{#1}}\fi
\expandafter\ifx\csname urlprefix\endcsname\relax\def\urlprefix{URL }\fi
\expandafter\ifx\csname href\endcsname\relax
  \def\href#1#2{#2} \def\path#1{#1}\fi

\bibitem{krizhevsky2017imagenet}
A.~Krizhevsky, I.~Sutskever, G.~E. Hinton, Imagenet classification with deep
  convolutional neural networks, Communications of the ACM 60~(6) (2017)
  84--90.

\bibitem{yolo}
J.~Redmon, S.~Divvala, R.~Girshick, A.~Farhadi, You only look once: Unified,
  real-time object detection, in: Proceedings of the IEEE conference on
  computer vision and pattern recognition, 2016, pp. 779--788.

\bibitem{FasterRcnn}
S.~Ren, K.~He, R.~Girshick, J.~Sun, Faster r-cnn: Towards real-time object
  detection with region proposal networks, Advances in neural information
  processing systems 28 (2015).

\bibitem{MAML}
C.~Finn, P.~Abbeel, S.~Levine, Model-agnostic meta-learning for fast adaptation
  of deep networks, in: International conference on machine learning, PMLR,
  2017, pp. 1126--1135.

\bibitem{optimization}
S.~Ravi, H.~Larochelle, Optimization as a model for few-shot learning, in:
  International conference on learning representations, 2017.

\bibitem{lee2019meta}
K.~Lee, S.~Maji, A.~Ravichandran, S.~Soatto, Meta-learning with differentiable
  convex optimization, in: Proceedings of the IEEE/CVF conference on computer
  vision and pattern recognition, 2019, pp. 10657--10665.

\bibitem{ProtoNet}
J.~Snell, K.~Swersky, R.~Zemel, Prototypical networks for few-shot learning,
  Advances in neural information processing systems 30 (2017).

\bibitem{vinyals2016matching}
O.~Vinyals, C.~Blundell, T.~Lillicrap, D.~Wierstra, et~al., Matching networks
  for one shot learning, Advances in neural information processing systems 29
  (2016).

\bibitem{DeepEMD}
C.~Zhang, Y.~Cai, G.~Lin, C.~Shen, Deepemd: Few-shot image classification with
  differentiable earth mover's distance and structured classifiers, in:
  Proceedings of the IEEE/CVF conference on computer vision and pattern
  recognition, 2020, pp. 12203--12213.

\bibitem{zhu2023light}
H.~Zhu, R.~Zhao, Z.~Gao, Q.~Tang, W.~Jiang, Light transformer learning
  embedding for few-shot classification with task-based enhancement, Applied
  Intelligence 53~(7) (2023) 7970--7987.

\bibitem{IDeMe-Net}
Z.~Chen, Y.~Fu, Y.-X. Wang, L.~Ma, W.~Liu, M.~Hebert, Image deformation
  meta-networks for one-shot learning, in: Proceedings of the IEEE/CVF
  conference on computer vision and pattern recognition, 2019, pp. 8680--8689.

\bibitem{DualTriNet}
Z.~Chen, Y.~Fu, Y.~Zhang, Y.-G. Jiang, X.~Xue, L.~Sigal, Multi-level semantic
  feature augmentation for one-shot learning, IEEE Transactions on Image
  Processing 28~(9) (2019) 4594--4605.

\bibitem{AFHN}
K.~Li, Y.~Zhang, K.~Li, Y.~Fu, Adversarial feature hallucination networks for
  few-shot learning, in: Proceedings of the IEEE/CVF Conference on Computer
  Vision and Pattern Recognition, 2020, pp. 13470--13479.

\bibitem{STVAE}
Y.~Zhang, S.~Huang, X.~Peng, D.~Yang, Semi-identical twins variational
  autoencoder for few-shot learning, IEEE Transactions on Neural Networks and
  Learning Systems (2023).

\bibitem{dhillon2019baseline}
G.~S. Dhillon, P.~Chaudhari, A.~Ravichandran, S.~Soatto, A baseline for
  few-shot image classification, arXiv preprint arXiv:1909.02729 (2019).

\bibitem{chencloser}
W.-Y. Chen, Y.-C. Liu, Z.~Kira, Y.-C.~F. Wang, J.-B. Huang, A closer look at
  few-shot classification, in: International Conference on Learning
  Representations.

\bibitem{RFS}
Y.~Tian, Y.~Wang, D.~Krishnan, J.~B. Tenenbaum, P.~Isola, Rethinking few-shot
  image classification: a good embedding is all you need?, in: Computer
  Vision--ECCV 2020: 16th European Conference, Glasgow, UK, August 23--28,
  2020, Proceedings, Part XIV 16, Springer, 2020, pp. 266--282.

\bibitem{IER}
M.~N. Rizve, S.~Khan, F.~S. Khan, M.~Shah, Exploring complementary strengths of
  invariant and equivariant representations for few-shot learning, in:
  Proceedings of the IEEE/CVF conference on computer vision and pattern
  recognition, 2021, pp. 10836--10846.

\bibitem{PAL}
J.~Ma, H.~Xie, G.~Han, S.-F. Chang, A.~Galstyan, W.~Abd-Almageed,
  Partner-assisted learning for few-shot image classification, in: Proceedings
  of the IEEE/CVF International Conference on Computer Vision, 2021, pp.
  10573--10582.

\bibitem{HandCrafted}
Y.~Zhang, S.~Huang, F.~Zhou, Generally boosting few-shot learning with
  handcrafted features, in: Proceedings of the 29th ACM International
  Conference on Multimedia, 2021, pp. 3143--3152.

\bibitem{Spatial}
Y.~Ouali, C.~Hudelot, M.~Tami, Spatial contrastive learning for few-shot
  classification, in: Machine Learning and Knowledge Discovery in Databases.
  Research Track: European Conference, ECML PKDD 2021, Bilbao, Spain, September
  13--17, 2021, Proceedings, Part I 21, Springer, 2021, pp. 671--686.

\bibitem{lee2020self}
H.~Lee, S.~J. Hwang, J.~Shin, Self-supervised label augmentation via input
  transformations, in: International Conference on Machine Learning, PMLR,
  2020, pp. 5714--5724.

\bibitem{IEPT}
M.~Zhang, J.~Zhang, Z.~Lu, T.~Xiang, M.~Ding, S.~Huang, Iept: Instance-level
  and episode-level pretext tasks for few-shot learning, in: International
  Conference on Learning Representations, 2021.

\bibitem{ESPT}
Y.~Rong, X.~Lu, Z.~Sun, Y.~Chen, S.~Xiong, Espt: A self-supervised episodic
  spatial pretext task for improving few-shot learning, in: Proceedings of the
  AAAI Conference on Artificial Intelligence, 2023.

\bibitem{SupCon}
P.~Khosla, P.~Teterwak, C.~Wang, A.~Sarna, Y.~Tian, P.~Isola, A.~Maschinot,
  C.~Liu, D.~Krishnan, Supervised contrastive learning, Advances in neural
  information processing systems 33 (2020) 18661--18673.

\bibitem{SimCLR}
T.~Chen, S.~Kornblith, M.~Norouzi, G.~Hinton, A simple framework for
  contrastive learning of visual representations, in: International conference
  on machine learning, PMLR, 2020, pp. 1597--1607.

\bibitem{MoCo}
K.~He, H.~Fan, Y.~Wu, S.~Xie, R.~Girshick, Momentum contrast for unsupervised
  visual representation learning, in: Proceedings of the IEEE/CVF conference on
  computer vision and pattern recognition, 2020, pp. 9729--9738.

\bibitem{ren2018meta}
M.~Ren, E.~Triantafillou, S.~Ravi, J.~Snell, K.~Swersky, J.~B. Tenenbaum,
  H.~Larochelle, R.~S. Zemel, Meta-learning for semi-supervised few-shot
  classification, arXiv preprint arXiv:1803.00676 (2018).

\bibitem{bertinetto2018meta}
L.~Bertinetto, J.~F. Henriques, P.~H. Torr, A.~Vedaldi, Meta-learning with
  differentiable closed-form solvers, arXiv preprint arXiv:1805.08136 (2018).

\bibitem{wah2011caltech}
C.~Wah, S.~Branson, P.~Welinder, P.~Perona, S.~Belongie, The caltech-ucsd
  birds-200-2011 dataset (2011).

\bibitem{guo2023contrastive}
H.~Guo, L.~Shi, Contrastive learning with semantic consistency constraint,
  Image and Vision Computing (2023) 104754.

\bibitem{wang2022arco}
Z.~Wang, S.~Shi, Z.~Zhai, Y.~Wu, R.~Yang, Arco: Attention-reinforced
  transformer with contrastive learning for image captioning, Image and Vision
  Computing 128 (2022) 104570.

\bibitem{zhao2023dual}
Y.~Zhao, Q.~Shu, X.~Shi, Dual-level contrastive learning for unsupervised
  person re-identification, Image and Vision Computing 129 (2023) 104607.

\bibitem{zhao2023unsupervised}
Y.~Zhao, Q.~Shu, X.~Shi, J.~Zhan, Unsupervised person re-identification by
  dynamic hybrid contrastive learning, Image and Vision Computing 137 (2023)
  104786.

\bibitem{SSLforFSL}
D.~Chen, Y.~Chen, Y.~Li, F.~Mao, Y.~He, H.~Xue, Self-supervised learning for
  few-shot image classification, in: ICASSP 2021-2021 IEEE International
  Conference on Acoustics, Speech and Signal Processing (ICASSP), IEEE, 2021,
  pp. 1745--1749.

\bibitem{JS1}
D.~M. Endres, J.~E. Schindelin, A new metric for probability distributions,
  IEEE Transactions on Information theory 49~(7) (2003) 1858--1860.

\bibitem{JS2}
B.~Fuglede, F.~Topsoe, Jensen-shannon divergence and hilbert space embedding,
  in: International symposium onInformation theory, 2004. ISIT 2004.
  Proceedings., IEEE, 2004, p.~31.

\bibitem{wu2018unsupervised}
Z.~Wu, Y.~Xiong, S.~X. Yu, D.~Lin, Unsupervised feature learning via
  non-parametric instance discrimination, in: Proceedings of the IEEE
  conference on computer vision and pattern recognition, 2018, pp. 3733--3742.

\bibitem{RENet}
D.~Kang, H.~Kwon, J.~Min, M.~Cho, Relational embedding for few-shot
  classification, in: Proceedings of the IEEE/CVF International Conference on
  Computer Vision, 2021, pp. 8822--8833.

\bibitem{AssoAlign}
A.~Afrasiyabi, J.-F. Lalonde, C.~Gagn{\'e}, Associative alignment for few-shot
  image classification, in: Computer Vision--ECCV 2020: 16th European
  Conference, Glasgow, UK, August 23--28, 2020, Proceedings, Part V 16,
  Springer, 2020, pp. 18--35.

\bibitem{GIFSL}
P.~Mazumder, P.~Singh, V.~P. Namboodiri, Gifsl-grafting based improved few-shot
  learning, Image and Vision Computing 104 (2020) 104006.

\bibitem{MELR}
N.~Fei, Z.~Lu, T.~Xiang, S.~Huang, Melr: Meta-learning via modeling
  episode-level relationships for few-shot learning, in: International
  Conference on Learning Representations, 2021.

\bibitem{PDA}
G.~Liu, L.~Zhao, X.~Fang, Pda: Proxy-based domain adaptation for few-shot image
  recognition, Image and Vision Computing 110 (2021) 104164.

\bibitem{HGNN}
T.~Yu, S.~He, Y.-Z. Song, T.~Xiang, Hybrid graph neural networks for few-shot
  learning, in: Proceedings of the AAAI conference on artificial intelligence,
  Vol.~36, 2022, pp. 3179--3187.

\bibitem{APP2S}
R.~Ma, P.~Fang, T.~Drummond, M.~Harandi, Adaptive poincar{\'e} point to set
  distance for few-shot classification, in: Proceedings of the AAAI conference
  on artificial intelligence, Vol.~36, 2022, pp. 1926--1934.

\bibitem{DGAP}
Z.~Cui, N.~Lu, W.~Wang, G.~Guo, Dual global-aware propagation for few-shot
  learning, Image and Vision Computing 128 (2022) 104574.

\bibitem{Meta-HP}
L.~Zhang, F.~Zhou, W.~Wei, Y.~Zhang, Meta-hallucinating prototype for few-shot
  learning promotion, Pattern Recognition 136 (2023) 109235.

\bibitem{SAPENet}
X.~Huang, S.~H. Choi, Sapenet: Self-attention based prototype enhancement
  network for few-shot learning, Pattern Recognition 135 (2023) 109170.

\bibitem{DFR}
H.~Cheng, Y.~Wang, H.~Li, A.~C. Kot, B.~Wen, Disentangled feature
  representation for few-shot image classification, IEEE Transactions on Neural
  Networks and Learning Systems (2023).

\bibitem{MIFN}
R.~Gao, H.~Su, S.~Prasad, P.~Tang, Few-shot classification with multisemantic
  information fusion network, Image and Vision Computing 141 (2024) 104869.

\bibitem{MetaDiff}
B.~Zhang, C.~Luo, D.~Yu, X.~Li, H.~Lin, Y.~Ye, B.~Zhang, Metadiff:
  Meta-learning with conditional diffusion for few-shot learning, in:
  Proceedings of the AAAI Conference on Artificial Intelligence, Vol.~38, 2024,
  pp. 16687--16695.

\bibitem{IbM2}
M.~Fu, K.~Zhu, Instance-based max-margin for practical few-shot recognition,
  in: Proceedings of the IEEE/CVF Conference on Computer Vision and Pattern
  Recognition, 2024, pp. 28674--28683.

\bibitem{FEAT}
H.-J. Ye, H.~Hu, D.-C. Zhan, F.~Sha, Few-shot learning via embedding adaptation
  with set-to-set functions, in: Proceedings of the IEEE/CVF conference on
  computer vision and pattern recognition, 2020, pp. 8808--8817.

\bibitem{MetaBaseline}
Y.~Chen, Z.~Liu, H.~Xu, T.~Darrell, X.~Wang, Meta-baseline: Exploring simple
  meta-learning for few-shot learning, in: Proceedings of the IEEE/CVF
  International Conference on Computer Vision, 2021, pp. 9062--9071.

\bibitem{ConstellationNet}
W.~Xu, Y.~Xu, H.~Wang, Z.~Tu, Attentional constellation nets for few-shot
  learning, in: International Conference on Learning Representations, 2021.

\end{thebibliography}

\end{document}